\pdfoutput=1

\documentclass[11pt]{article}

\usepackage[]{acl}

\usepackage{times}
\usepackage{latexsym}

\usepackage[T1]{fontenc}

\usepackage[utf8]{inputenc}

\usepackage{microtype}

\usepackage{inconsolata}

\usepackage{amsmath,graphicx}
\usepackage{multirow}
\usepackage{xtab,booktabs}
\usepackage{longtable}
\usepackage{bm}
\usepackage{amssymb}
\usepackage{caption}
\usepackage{subcaption}
\usepackage{xcolor}
\usepackage{hyperref}

\title{Teaching a Multilingual Large Language Model to Understand Multilingual Speech via Multi-Instructional Training}
\author{
Pavel Denisov \and Ngoc Thang Vu\\
University of Stuttgart, Institute for Natural Language Processing, Germany\\
\texttt{\{pavel.denisov, thang.vu\}@ims.uni-stuttgart.de}
}
\begin{document}
\maketitle
\begin{abstract}
Recent advancements in language modeling have led to the emergence
of Large Language Models (LLMs) capable of
various natural language processing tasks.
Despite their success in text-based tasks, applying LLMs to the speech domain
remains limited and challenging. This paper presents BLOOMZMMS, a novel model
that integrates a multilingual LLM with a multilingual speech encoder,
aiming to harness the capabilities of LLMs for speech recognition and beyond.
Utilizing a multi-instructional training approach, we demonstrate the transferability
of linguistic knowledge from the text to the speech modality.
Our experiments, conducted on 1900 hours of transcribed data from 139 languages,
establish that a multilingual speech representation can be effectively
learned and aligned with a multilingual LLM. While this learned representation
initially shows limitations in task generalization, we address this issue by
generating synthetic targets in a multi-instructional style.
Our zero-shot evaluation results confirm the robustness of our approach across
multiple tasks, including speech translation and multilingual spoken language
understanding, thereby opening new avenues for applying LLMs in the speech domain.
\end{abstract}

\section{Introduction}

Language modeling task involves predicting subsequent text tokens based
on a context of preceding ones \cite{jurafsky2009speech}. 
Training a language model (LM) requires only raw text samples, as portions of these samples function as their labels, facilitating a self-supervised learning (SSL) approach.  The widespread availability of machine-readable text online, coupled with advancements in computational power, has led to the rise of large LMs (LLMs) in recent years.
These LLMs not only generate highly fluent natural text but also encode higher-level
knowledge within their parameters. This enables them to tackle natural language
processing tasks like reading comprehension and machine translation
based only on task specific instructions, without needing labeled data \cite{radford2019language}.

SSL has recently made significant strides in the speech domain \cite{baevski2020wav2vec}.
Most applications of SSL in speech employ an encoder that transforms
raw speech signals into high-level representations, serving either
as a fixed feature extractor \cite{yang2021superb} or a tunable pretrained model
for various downstream tasks \cite{babu2021xls}. Incorporating of SSL pretrained encoders
into Encoder-Decoder speech recognition models has dramatically reduced
the amount of labeled data required for effective training \cite{chang2021exploration}.
However, using SSL pretrained decoders in such models is relatively rare.
In certain instances, SSL is part of a joint training process that seeks
to learn a shared speech and text representation \cite{chen2022maestro}. However, this approach often
demands a large dataset and considerable computational resources.

Recent work has begun to harness the powerful text generation capabilities
of decoder-only LLMs by incorporating them as the decoder component of Encoder-Decoder
speech processing models.
\citet{wu2023decoder} adopt the LLaMA-7B LLM for speech translation to English
by training a speech encoder from scratch using filter bank acoustic features,
14,000 hours of internal speech data in 14 languages, and outputs of internal
translation system as synthetic targets. Outputs of speech encoder are aligned with
the text token embedding space using CTC pretraining and downsampled by
averaging of consequative frames with the same CTC output label.
\citet{ling2023adapting} adopt the GPT2 XL LLM for fully-formatted English speech recognition
by training a speech encoder from scratch using filter bank acoustic features,
and 75,000 hours of internal transcribed English speech data. CTC loss is applied to speech encoder
outputs as a part of the main training process and speech representations are downsampled
by removal of frames classified as CTC blank labels with a predefined threshold.
\citet{li2023prompting} adopt the LLaMA-7B LLM for long-form English speech recognition
by incorporating the HuBERT-Large SSL pretrained speech encoder and finetuining
it on the LibriSpeech dataset containing 960 hours of transcribed English speech.
Outputs of the speech encoder are downsampled by a convolutional module trained
as a part of the main training process.
\citet{fathullah2023prompting} adopt the LLaMA-7B LLM for speech recognition in 8 languages
by training a speech encoder from scratch using filter bank acoustic features
and the Multilingual LibriSpeech dataset containing 50,000 hours of transcribed speech in the same 8 languages.
Speech encoder is pretrained with CTC loss and its outputs are downsampled by simple
discarding of every $n$ frames.
\citet{nachmani2023spoken} combine an internal pretrained LLM with an internal pretrained speech encoder
and finetune it on the automatically transcribed LibriLight dataset containing 60,000 hours of English speech.
The training is performed with a combination of the speech transcription and speech continuation tasks.
The resulting model is utilized for the spoken language answering task.
Most of these studies rely on conventional
filter bank features for speech encoding and do not incorporate an SSL pretrained speech encoder,
necessitating a large amount of training data. Moreover, scant attention has been
given to leveraging the linguistic knowledge stored in LLMs for tasks beyond mere transcription
and for languages other than English.

To address these challenges, we propose BLOOMZMMS, a model that fuses a multilingual
LLM (BLOOMZ \cite{muennighoff-etal-2023-crosslingual})
with a multilingual speech encoder (MMS \cite{pratap2023scaling}).
We argue that multi-instructional training is
crucial for transferring linguistic knowledge from the text to speech modality.
Our experiments demonstrate that training
on 1900 hours of transcribed data from 139 languages
yields a multilingual speech representation compatible with a multilingual
LLM in the context of Automatic Speech Recognition (ASR) task.
Although this representation does not generalize well to other tasks,
we show that the issue can be mitigated by generating additional synthetic targets.
Our zero-shot evaluations confirm this approach's effectiveness across various tasks,
including Spoken Language Translation (SLT) and multilingual spoken Natural Language Inference (NLI).
Our training recipes and models are released under
the Apache-2.0 license\footnote{\url{https://github.com/akreal/bloomzmms}}.

\section{Method}

\begin{figure*}[htb]
	\begin{subfigure}[b]{.6\linewidth}
		\centerline{\includegraphics[scale=0.6]{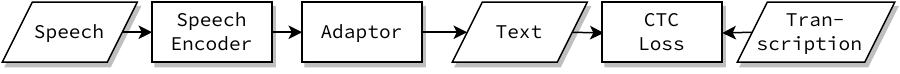}}
		\caption{Stage one with the CTC training.}
	\end{subfigure}

	\begin{subfigure}[b]{.6\linewidth}
		\vspace{-30pt}
		\centerline{\includegraphics[scale=0.6]{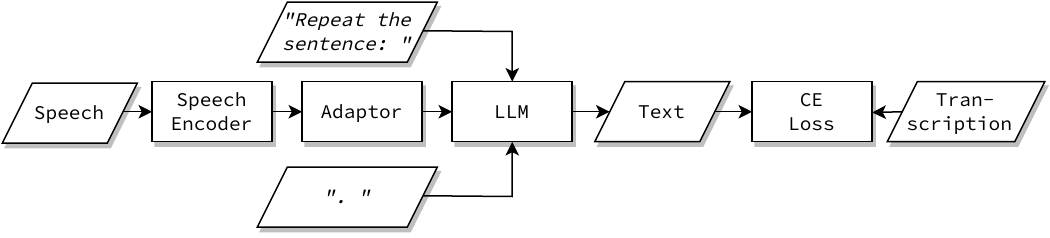}}
		\caption{Stage two with the transcription task training (T).}
	\end{subfigure}%
	\begin{subfigure}[b]{.5\linewidth}
		\vspace{-50pt}
		\centerline{\includegraphics[scale=0.6]{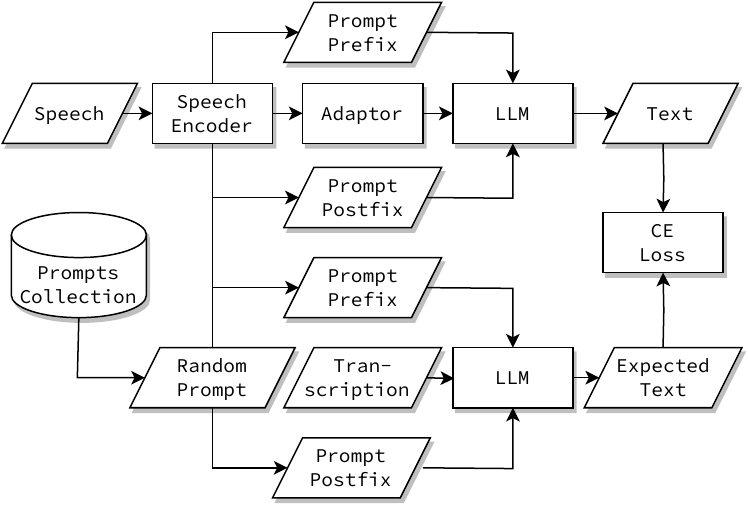}}
		\caption{Stage two with the multi-instructional training (MI).}
	\end{subfigure}
	\caption{
	Overview of the Adaptor training.
	At the stage one, the Adaptor parameters are optimized using the CTC loss to directly predict the transcription (a).
	At the stage two, the Adaptor parameters are optimized using the CE loss applied to the outputs of the LLM while
	the Adaptor output is enclosed in the prompt's prefix and postfix text and is fed to the LLM input.
	A prompt can instruct the model to generate a transcription (b) or perform some other task
	on the speech input (c). In the case of transcription, a ground truth transcription is used as a training target.
	In the case of other instructions, a training target is obtained by running the LLM inference with
	the same prompt and ground truth transcription as the input.
	}
	\label{fig:method}
\end{figure*}

The proposed method is outlined in Figure \ref{fig:method}.
Our model comprises the pretrained speech encoder, LLM and an intermediate
Adaptor module that maps the output of the speech encoder to the latent space
of the text token embeddings of the LLM.
We train the Adaptor module using pairs of speech recordings and their corresponding text transcriptions,
denoted as $\bm{x}$ and $\bm{y}^{\rm Transcription}$ respectively, and keep the parameters
of the speech encoder and the LLM frozen. The objective of the Adaptor training
is to make its output $\bm{H}^{\rm Adaptor}$ obtained from the input speech $\bm{x}$
as close as possible to the text embedding sequence
of the ground truth transcription $\bm{H}^{\rm Transcription} = \text{LMEmbedding}(\bm{y}^{\rm Transcription})$,
where $\text{LMEmbedding}$ is the token embedding layer of the LLM.

Similarly to previous works on the LLM adaptation to the speech modality \cite{wu2023decoder,fathullah2023prompting},
our training process comprises of the two stages:
an alignment of the speech encoder output with the LLM token embedding space,
and an integrated optimization of the complete model with the LLM.
An attempt to omit either of the two stages in our process
leads to the lack of training convergence.
We hypothesize that the different training stages help the Adaptor to learn
different subtasks like segmentation, ordering and the actual token embedding prediction.

At the first stage of the training, $\bm{H}^{\rm Adaptor}$ is projected
to the LLM tokens' logits using the frozen output linear layer of the LLM
(which is often a transposed token embedding layer),
and the Connectionist Temporal Classification (CTC) loss \cite{graves2006connectionist}
is minimized between the LLM token probabilities obtained
from the token logits and the transcription:

\begin{footnotesize}
\begin{align*}
\bm{H}^{\rm Speech} &= \text{SpeechEncoder}(\bm{x}) \\
\bm{H}^{\rm Adaptor} &= \text{Adaptor}(\bm{H}^{\rm Speech}) \\
p_{\rm CTC}(\bm{y}|\bm{x}) &= \text{Softmax}(\bm{H}^{\rm Adaptor} \bm{W}) \\
\mathcal{L}_{\rm CTC} & = - \sum_{\bm{\pi} \in \mathcal{B}^{-1}(\bm{y}^{\rm Transcription})} \log p_{\rm CTC}(\bm{\pi} | \bm{x}),
\end{align*}
\end{footnotesize}
where the mapping $\mathcal{B}$ removes
repeated and blank tokens according to the CTC definition,
$\bm{W} \in \mathbb{R}^{d \times v}$ is the transposed weight matrix of the token embedding layer,
$d$ is the dimensionality of the embedding, and $v$ is the number of tokens in the LLM's vocabulary.

At the second stage, $\bm{H}^{\rm Adaptor}$ is concatenated with
the token embeddings of the prefix and postfix parts of a text prompt.
This joint sequence is then passed through the self-attention layers
of the LLM and  projected with the transposed token embedding weight matrix $W$
(also serving as the output layer of the LLM) to obtain the LLM prediction.
The Cross-Entropy (CE) loss is minimized between the prediction of the LLM
for this sequence and the expected LLM output.
In case of the speech recognition task,
we set the prompt prefix and postfix to
\texttt{"Repeat~the~sentence:~"} and \texttt{".~"} respectively:

\begin{footnotesize}
\begin{align*}
\bm{H}^{\rm Prefix} &= \text{LMEmbedding}(\small{\texttt{"Repeat the sentence: "}}) \\
\bm{H}^{\rm Postfix} &= \text{LMEmbedding}(\small{\texttt{". "}}) \\
\bm{H}^{\rm LM} &= \text{LM}((\bm{H}^{\rm Prefix}, \bm{H}^{\rm Adaptor}, \bm{H}^{\rm Postfix})) \\
p_{\rm CE}(\bm{y}|\bm{x}) &= \text{Softmax}(\bm{H}^{\rm LM} \bm{W}) \\
\mathcal{L}_{\rm CE-ASR} & = - \log p_{\rm CE}(\bm{y}^{\rm Transcription}|\bm{x}),
\end{align*}
\end{footnotesize}
where $\text{LM}()$ denotes the self-attention layers of the LLM.
In case of the multi-instructional training, prompts are sampled
from a predefined hand crafted collection, while the expected
output is set to the output of the LLM for the same prompt
using the token embeddings of the ground truth
transcription instead of the Adaptor output $\bm{H}^{\rm Adaptor}$:

\begin{footnotesize}
\begin{align*}
\bm{H}^{\rm Prefix} &= \text{LMEmbedding}(\bm{p}^{i}_{\rm Prefix}) \\
\bm{H}^{\rm Postfix} &= \text{LMEmbedding}(\bm{p}^{i}_{\rm Postfix}) \\
\bm{H}^{\rm LM} &= \text{LM}((\bm{H}^{\rm Prefix}, \bm{H}^{\rm Adaptor}, \bm{H}^{\rm Postfix})) \\
\bm{H}^{\rm LM-Text} &= \text{LM}((\bm{H}^{\rm Prefix}, \bm{H}^{\rm Transcription}, \bm{H}^{\rm Postfix})) \\
\bm{y}^{\rm LM} &= \text{BeamSearch}(\text{Softmax}(\bm{H}^{\rm LM-Text} \bm{W})) \\
p_{\rm CE}(\bm{y}|\bm{x}) &= \text{Softmax}(\bm{H}^{\rm LM} \bm{W}) \\
\mathcal{L}_{\rm CE-MI} & = - \log p_{\rm CE}(\bm{y}^{\rm LM}|\bm{x}),
\end{align*}
\end{footnotesize}
where $\bm{p}^{i}_{\rm Prefix}$ and $\bm{p}^{i}_{\rm Postfix}$ are the prefix and postfix
texts of the $i$-th prompt in the prompts collection,
$i \sim U([1, \dots , N_{\rm Pr}])$ is a random number drawn from an uniform distribution
over all natural numbers between 1 and $N_{\rm Pr}$,
and $N_{\rm Pr}$ is the number of prompts in the collection.

\section{Experiments}

\subsection{Training and Validation Data}

The Adaptor training is performed on the entire training FLEURS dataset \cite{conneau2023fleurs}
and a subset of the Common Voice Corpus 12.0 \cite{ardila-etal-2020-common} training dataset
with the total amount of 993,660 utterances or 1905 hours of recordings.
The Common Voice subset is constructed by selection of up to 25 hours
of recordings for each language.
Our validation set is the validation set of FLEURS
with the total amount of 34,044 utterances or 115 hours of recordings.
All transcriptions are taken in an unnormalized format with the true casing
and punctuation.

Multi-instructional training labels are synthesized with prompts
from the P3 collection \cite{sanh2022multitask}.
The P3 collection is selected because it was employed in the finetuning
process of transitioning BLOOM into BLOOMZ.
Our objective is to ensure consistent output for both speech and text inputs.
To achieve this, we generate text outputs utilizing prompts
from the P3 collection, with which the BLOOMZ model is already acquainted.
We apply six distinct randomly drawn prompts to
a transcription of each original utterance and assign two generated
outputs to each of the three speed-perturbed versions of that utterance.
The outputs are generated with a greedy search and maximum length of 128 tokens.

\subsection{Evaluation Data and Metrics}

We evaluate our model on the following established benchmarks:
FLEURS \cite{conneau2023fleurs}, MLS \cite{pratap2020mls}
and VoxPopuli \cite{wang-etal-2021-voxpopuli} for the ASR, CoVoST~2 \cite{wang2021covost} for the SLT,
SpeechGLUE \cite{ashihara2023speechglue} for the spoken General Language Understanding (GLUE) and
SpeechXNLI for the multilingual NLI\footnote{Following SpeechGLUE,
we synthesize a speech version of the XNLI \cite{conneau-etal-2018-xnli} validation subset
using the IMS Toucan \cite{lux-etal-2022-low} text-to-speech toolkit: \url{https://zenodo.org/records/10900287}.}.
The results are evaluated using the corresponding metrics:
Word Error Rate (WER) and Character Error Rate (CER) for the ASR,
BLEU\footnote{Using the SacreBLEU tool \cite{post-2018-call}.} \cite{papineni-etal-2002-bleu} for the SLT,
Matthews Correlation Coefficient (MCC) for the CoLA task within SpeechGLUE,
and accuracy for the other SpeechGLUE tasks and the SpeechXNLI.
Whisper normalization
is applied for both reference and hypothesis before evaluating CER/WER in the ASR experiments.

\subsection{Experimental Setup}

Our model is implemented using ESPnet2 \cite{watanabe20212020} version 202304
and Hugging Face Transformers \cite{wolf-etal-2020-transformers} version 4.31.0.
We use weighted-sum of hidden states \cite{yang2021superb,chang2021exploration}
of the MMS 1B-ASR-All\footnote{\url{https://huggingface.co/facebook/mms-1b-all}}
pretrained model \cite{pratap2023scaling} as speech features.
We discard all language specific adapters and heads of the MMS 1B-ASR-All model
to simplify the implementation while preserving the multilingual properties of our system.
The Adaptor module is a VGG/E-Branchformer based encoder \cite{kim2023branchformer}
combined with a convolutional Length Adaptor \cite{li-etal-2021-multilingual}.
The E-Branchformer encoder is configured with 17 layers,
each with 2048 hidden units, 8 attention heads, and output dimension of 1024.
The Convolutions to Gated MultiLayer Perceptron module has 8192 units and the convolution kernel size is 31.
The Length Adaptor module contains a 1-dimensional convolutional layer with stride 2
and reduces the length of input sequence by factor of 2.
Self-conditioning on language identity \cite{chen2023improving}
is applied during the CTC training.
The LLM in our experiments is BLOOMZ 7.1B\footnote{\url{https://huggingface.co/bigscience/bloomz-7b1}}
model \cite{muennighoff-etal-2023-crosslingual},
which itself is BLOOM 7.1B LLM \cite{workshop2023bloom} finetuned on
the xP3 dataset introduced with BLOOMZ.
The total number of parameters in our model is 8.6 billions,
the number of trainable parameters is 536 millions.
We apply 8-bit quantization \cite{dettmers2022llm} to the LLM
using the functions from the \texttt{bitsandbytes} package version 0.41.1.
The training is done with the Adam optimizer \cite{adam} with
$\beta_1=0.9$, $\beta_2=0.999$, $\epsilon=10^{-8}$,
the warmup learning rate scheduler with
the maximum learning rate of $10^{-4}$ and a weight decay of $10^{-6}$.
3-way speed perturbation \cite{ko2015audio} data augmentation
method is applied to the training data.

The training stage one, \textbf{CTC loss training}, is performed on two NVIDIA RTX A6000 GPUs
with the global batch size of 7.29 minutes.
The number of warmup steps for the learning rate scheduler is set to 25,000.
A checkpoint is saved every 23,364 steps and evaluated on the validation dataset.
The training is stopped after four consecutive evaluations showing no improvement,
it takes 233,640 update steps or 120 hours of training time to reach this condition.
A checkpoint with the lowest validation CER from the stage one is used
to initialize the model for the stage two.

The training stage two, \textbf{CE loss training}, is performed on four NVIDIA RTX A6000 GPUs
with the batch size of 37.50 seconds and a gradient accumulation over two batches.
The number of warmup steps for the learning rate scheduler is set to 10,000.
A checkpoint is saved every 54,381 steps and evaluated on the validation dataset.
The training is stopped after four consecutive evaluations showing no improvement.
To reach this condition, it takes
652,572 update steps or 132 hours of training on the transcription targets,
2,664,669 update steps or 686 hours on the multi-instructional targets, and
2,501,526 update steps or 644 hours on the combined set of targets.
A checkpoint with the highest validation token prediction accuracy from the second step
is used for the zero-shot evaluations.

We decode with the beam search of size 5 and set the maximum output sequence to 192 tokens
to obtain the model predictions for the ASR and SLT evaluations.
The GLUE and NLI evaluations restrict the output to the possible answer options
corresponding to a task and limits the beam size and maximum output sequence
respectively. For example, for a yes/no question the possible outputs are
\texttt{yes} or \texttt{no}, the beam size is 2 and the maximum output sequence is 1.
All evaluations are executed on one NVIDIA RTX A6000 GPU.

\section{Results}

\subsection{Multitasking}

\begin{table}[!ht]
    \center
    {
    \footnotesize
    \setlength{\tabcolsep}{1.0mm}
    \begin{tabular*}{1.0\columnwidth}{l  l  l  r r  r}
    \toprule
    \multirow{2}{*}{Task}  & \multirow{2}{*}{Dataset}  & \multirow{2}{*}{Metrics} & \multicolumn{3}{c}{Training targets}  \\
    \cmidrule{4-6}
      &   &  & \multicolumn{1}{c}{T} & \multicolumn{1}{c}{MI}  & TMI \\
    \midrule
     ASR & FLEURS & CER$\downarrow$          & \textbf{12.0} & 88.5 & 12.4 \\
     SLT & CoVoST~2 X$\rightarrow$En & BLEU$\uparrow$          & 3.0 &  14.1 & \textbf{15.6} \\
     GLUE & SpeechGLUE & Acc./MCC $\uparrow$ & 41.7 & 54.4 & \textbf{55.9} \\
     NLI & SpeechXNLI & Acc. $\uparrow$ & 35.8 & \textbf{41.6}  & 41.4 \\
    \bottomrule
  \end{tabular*}
  }
    \caption{
    	Comparative performance metrics across various speech processing tasks
	using different training targets: transcription (T), synthetic multi-instructional (MI)
	and their combination (TMI).}
    \label{tab:multitask_results}
\end{table}

Table \ref{tab:multitask_results} presents evaluation results
of our model across various speech processing tasks,
including multilingual ASR, SLT, spoken GLUE, and multilingual NLI.
These evaluations test three versions of the model, which are trained using
different training targets: transcription only (T),
Multi-Instruction (MI), and a combination of both (TMI).
When the model is trained solely on the transcription task,
it achieves good performance for the ASR task itself, with a CER of 12.0.
However, this specialized training does not generalize well
to more sophisticated tasks like SLT, GLUE, or NLI,
as evidenced by the notably lower performance metrics.
On the other hand, training the model on MI
synthetic targets shows significant improvement in performing
other tasks such as SLT, GLUE, and NLI. The BLEU score for SLT,
for example, increases to 14.1 and the average accuracy/MCC score for GLUE rises to 54.4.
Despite these gains, the MI-only training leads to a significant drop in
performance for the ASR task, registering a CER of 88.5.
Combining both transcription and MI
targets enables the model to perform well across all tested tasks.
In addition to maintaining strong performance in ASR (CER of 12.4),
this training configuration also leads to improvements in two out of the three
non-ASR tasks.
These results underscore the benefits of integrating ASR and
MI targets.

\subsection{Speech Recognition}

\begin{table}[!ht]
    \center
    {
    \footnotesize
    \setlength{\tabcolsep}{1.5mm}
    \begin{tabular*}{.92\columnwidth}{l l r r r}
    \toprule
    \multirow{2}{*}{Dataset}  & \multirow{2}{*}{Languages} & \multicolumn{3}{c}{Training targets}  \\
    \cmidrule{3-5}
      &    & \multicolumn{1}{c}{T} & \multicolumn{1}{c}{MI}  & TMI  \\
    \midrule
    \multirow{3}{1.2cm}{FLEURS (CER)}  & BLOOM (34)     & 12.9 & 70.0 & \textbf{12.8} \\
                                       & Non-BLOOM (68) & \textbf{11.6} & 86.4 & 12.2 \\
                                       & All (102)      & \textbf{12.0} & 80.9 & 12.4 \\
    \midrule
    \multirow{3}{1.2cm}{MLS}                      & BLOOM (4)     & 27.0 & 25.0 & \textbf{20.6} \\
                                                  & Non-BLOOM (4) & \textbf{11.5} & 72.4 & 12.2 \\
                                                  & All (8)       & 19.3 & 48.7 & \textbf{16.4} \\
    \midrule
    \multirow{3}{*}{VoxPopuli}  & BLOOM (3)      & 21.3 & 22.0 & \textbf{17.3} \\
                                & Non-BLOOM (11) & 22.2 & 104.6 & \textbf{22.0} \\
                                & All (14)       & 22.0 & 86.9 & \textbf{21.0} \\
    \bottomrule
    \end{tabular*}
    }
    \caption{
	Comparative evaluation of speech recognition performance depending on the training targets.
	Results are stratified by language exposure during BLOOM training and evaluated using WER,
	except for the FLEURS dataset that uses CER for compatibility with previous works.
	}
    \label{tab:results_recognition}
\end{table}

\begin{figure}[]
        \center
	\includegraphics[width=\columnwidth]{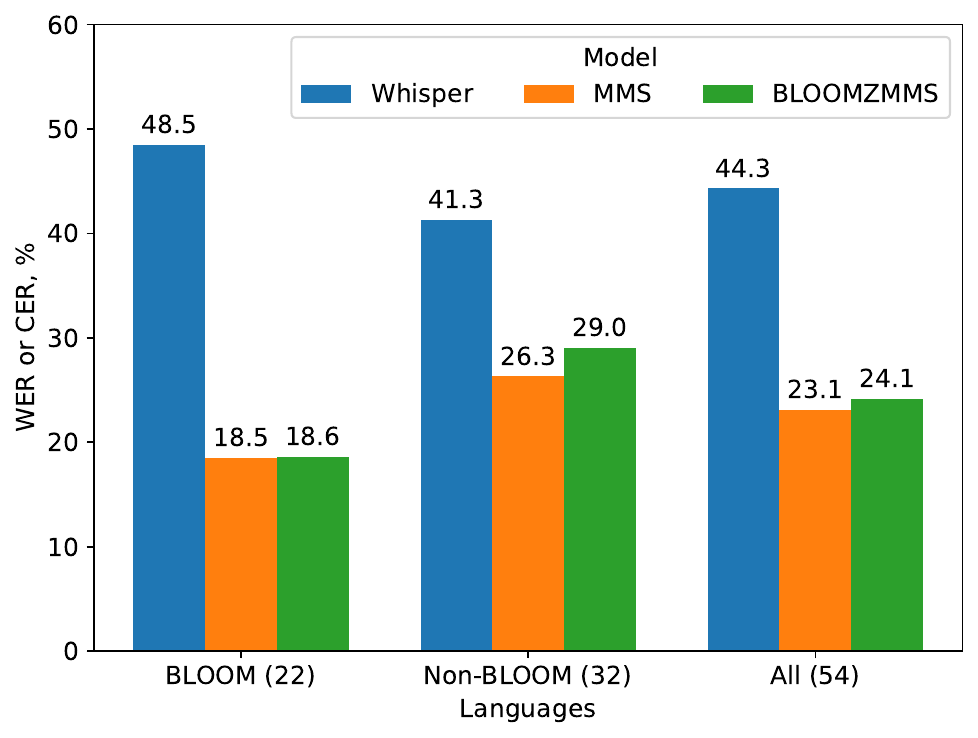}
	\caption{
	Comparative evaluation of speech recognition performance between the BLOOMZMMS TMI model
	and previous works, multi-domain MMS (1B) \cite{pratap2023scaling}
	and Whisper large-v2 \cite{radford2023robust}, on the FLEURS-54 evaluation dataset.
	All numbers are WER except for Thai, Lao, Burmese, and Khmer languages.
	}
	\label{fig:fleurs54}
\end{figure}

Table \ref{tab:results_recognition} presents a comparative analysis
of ASR performance for the BLOOMZMMS model
with the T, MI and TMI training targets.
Results are further divided based on whether the languages were seen during the training
of the BLOOM model or not.
For languages that were part of the BLOOM model training,
the TMI model generally performs better than the T model.
The opposite is true for the non-BLOOM languages.
This is expected as training on the MI targets puts stronger stress on the distillation
of the LLM knowledge and its encoding to the Adaptor parameters.
This effect is more pronounced on the MLS and VoxPopuli datasets,
which represent recording conditions and
linguistic content slightly different from our training data.
Nevertheless, both T and TMI BLOOMZMMS models perform comparably
on the in-domain FLEURS dataset independently from the language,
suggesting that the Adaptor can effectively leverage
the outputs of the MMS speech encoder in order to compensate
for the lack of language familiarity by the LLM.

Following the MMS paper, we separate a subset of FLEURS testing dataset
for the 54 languages that are supported by the Whisper model,
and compare the results of the BLOOMZMMS TMI model
to the results of the multi-domain MMS (1B) and Whisper large-v2 models.
The MMS model is essentially the same speech encoder as used by BLOOMZMMS,
but with a number of language-specific components,
namely adapter parameters, output vocabulary,
and n-gram model utilized during decoding.
Despite removal of the language-specific components
and addition of the other speech processing tasks,
such as SLT, BLOOMZMMS manages to keep the ASR
performance on a comparable level to the original MMS model.
While also being a multitask model, BLOOMZMMS
outperforms the other strong multitask alternative,
Whisper large-v2, by a large margin
on this massively multilingual low-resource ASR benchmark,
albeit potentially due to being trained on in-domain data,
in contrast to Whisper.

\subsection{Speech Translation}

\begin{table}[!ht]
    \center
    {
    \footnotesize
    \setlength{\tabcolsep}{1.5mm}
    \begin{tabular*}{0.97\columnwidth}{l l r r r r}
    \toprule
    \multirow{2}{*}{Dataset}  & \multirow{2}{*}{Languages} & \multicolumn{3}{c}{Training targets} & \multirow{2}{*}{\emph{Gold}}  \\
    \cmidrule{3-5}
      &  & \multicolumn{1}{c}{T}  & \multicolumn{1}{c}{MI} & TMI &   \\
    \midrule
    \multirow{3}{*}{X$\rightarrow$En}  & BLOOM (8)      & 7.0 & 25.9 & \textbf{26.8} & \emph{35.5} \\
                                       & Non-BLOOM (13) & 0.6 &  8.4 & \textbf{8.7}  & \emph{11.3} \\
                                       & All (21)       & 3.0 & 15.1 & \textbf{15.6} & \emph{20.5} \\
    \midrule
    \multirow{3}{*}{En$\rightarrow$X}  & BLOOM (5)      & 1.1 & 10.9 & \textbf{11.0} & \emph{17.5} \\
                                       & Non-BLOOM (10) & 0.3 &  0.9 & \textbf{1.0}  & \emph{1.7} \\
                                       & All (15)       & 0.5 &  4.2 & \textbf{4.3}  & \emph{7.0} \\
    \bottomrule
  \end{tabular*}
  }
    \caption{
    Comparative evaluation of zero-shot speech translation performance
    depending on the training targets using the CoVoST~2 dataset.
    Results are stratified by language exposure during
    BLOOM training and evaluated using BLEU metrics\protect\footnotemark.
    Results on text inputs (\emph{Gold}) are given for comparison.
	}
    \label{tab:results_translation_covost}
\end{table}
\footnotetext{sacreBLEU signature: nrefs:1 | case:mixed | eff:no | tok:13a | smooth:exp | version:2.3.1.}

\begin{figure}[]
        \center
	\includegraphics[width=\columnwidth]{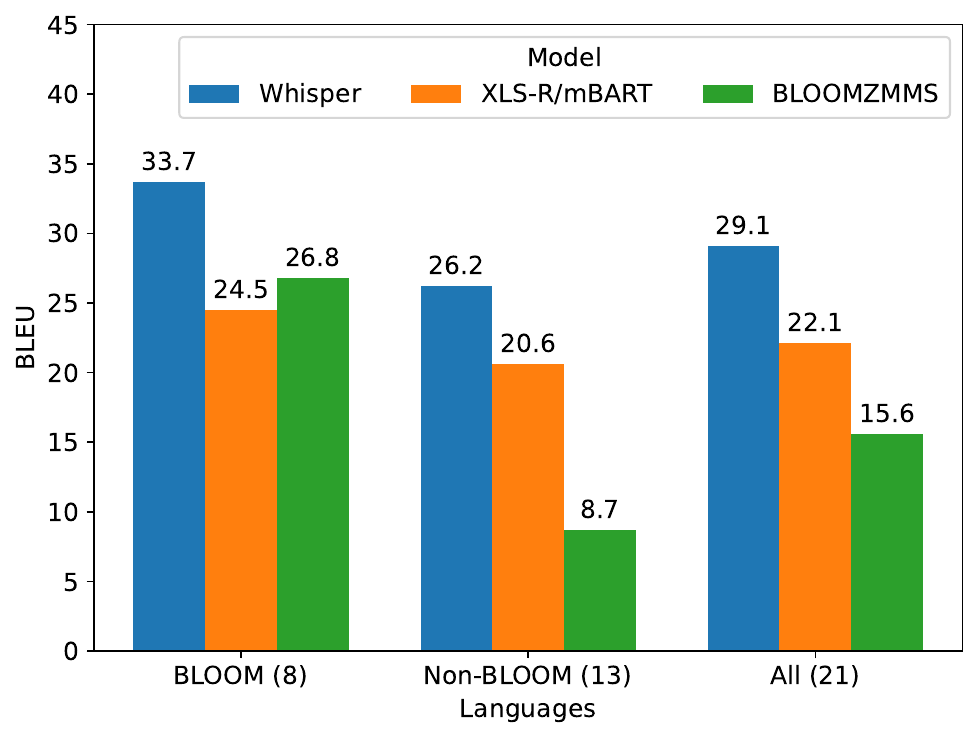}
	\caption{
	Comparative evaluation of speech translation performance between the BLOOMZMMS TMI model
	and previous works, XLS-R/mBART \cite{babu2021xls}
	and Whisper large-v2 \cite{radford2023robust},
	on the CoVoST~2 X$\rightarrow$En evaluation set.
	}
	\label{fig:covost2}
\end{figure}

Table \ref{tab:results_translation_covost} presents the zero-shot evaluation
results for SLT using the CoVoST~2 dataset.
The BLOOMZ LM exhibits a nascent ability
to translate languages that it has not been trained on,
and when this knowledge is transferred to the speech modality,
there's only a minor loss in accuracy.
Interestingly, the performance gap between the BLOOMZMMS model
and gold transcriptions is more pronounced for the BLOOM languages.
This indicates that the quality of knowledge transfer
from text to speech depends on the initial linguistic knowledge in the text-based LLM.
Consequently, weaknesses present in the LLM
tend to amplify when transferred to the speech modality,
suggesting that the proposed method might benefit from some form of
regularization to mitigate this effect.

Figure \ref{fig:covost2} shows the comparison of the BLOOMZMMS TMI
model with the previous works, XLS-R/mBART and Whisper large-v2,
for the X$\rightarrow$En translation direction.
XLS-R/mBART is a strong baseline, which is finetuned
on complete CoVoST~2 training data.
Whisper large-v2 has not seen any CoVoST~2 data during training,
but has been supervised by a large amount
of other speech translation data.
BLOOMZMMS TMI has not been exposed to any gold labeled
speech translation samples during training.
Remarkably, the zero-shot BLOOMZMMS model outperforms
the supervised task-specific XLS-R/mBART model
for the languages previously seen during BLOOM training.
This impressive result is primarily due to the strong performance
of the BLOOMZ LLM, which is successfully
transferred to the speech modality via the multi-instructional training.
However, there is a notable gap with the multitask
Whisper large-v2 model, primarily attributed to the
poor performance on unseen languages of the LLM we utilize.

\begin{table}[!ht]
    \center
    {
    \footnotesize
    \setlength{\tabcolsep}{1.5mm}
    \begin{tabular*}{0.97\columnwidth}{l l  r  r r r}
    \toprule
    \multirow{2}{*}{Dataset}  & \multirow{2}{*}{Languages} & \multicolumn{3}{c}{Training targets} & \multirow{2}{*}{\emph{Gold}}  \\
    \cmidrule{3-5}
      &  & \multicolumn{1}{c}{T} & \multicolumn{1}{c}{MI} & TMI &   \\
    \midrule
    \multirow{3}{*}{X$\rightarrow$En}  & BLOOM (33)      & 1.2  & 30.6          & \textbf{30.8} & \emph{44.4} \\
                                       & Non-BLOOM (67)  & 0.5  &  \textbf{8.6} &  8.3 & \emph{12.5} \\
                                       & All (100)       & 0.7  & \textbf{15.9} & 15.7 & \emph{23.1} \\
    \midrule
    \multirow{3}{*}{En$\rightarrow$X}  & BLOOM (33)      & 18.1 & 24.7 & \textbf{24.8} & \emph{30.0} \\
                                       & Non-BLOOM (67)  & 1.1  &  \textbf{1.2} &  \textbf{1.2} & \emph{1.9} \\
                                       & All (100)       & 6.7  &  8.9 &  \textbf{9.0} & \emph{11.2} \\
    \bottomrule
  \end{tabular*}
  }
    \caption{
    Comparative evaluation of zero-shot speech translation performance
    depending on the training targets using the FLEURS dataset.
    Results are stratified by language exposure during
    BLOOM training and evaluated using BLEU metrics.
    Results on text inputs (\emph{Gold}) are given for comparison.
	}
    \label{tab:results_translation_fleurs}
\end{table}

In order to expand language coverage, we evaluate our model for the SLT performance
on the FLEURS dataset as well, and present the results in
Table~\ref{tab:results_translation_fleurs}.
As suggested by \citet{radford2023robust}, we use target language transcriptions
for the sentences with the same ID as reference translations.
Our evaluation does not include Afrikaans, because
the version of the dataset we use\footnote{\url{https://huggingface.co/datasets/google/fleurs}}
does not include any sentence IDs shared between Afrikaans and English.
The multilingual properties of the BLOOMZ model, which serves as a decoder
of our model, enable us to report the SLT results with non-English target languages
as well, for the first time on the FLEURS dataset to the best of our knowledge.
The results confirm the good transferability of translation capabilities
from text to speech modality with the MI and TMI training targets
for a wider range of languages seen in the BLOOM training data.
The fair translation performance from unseen languages to English, as observed
in the CoVoST~2 dataset, can also be seen across a wider range
of languages in the FLEURS dataset.

\subsection{Spoken Language Understanding}

\begin{table}[]
    \center
    {
    \footnotesize
    \begin{tabular*}{0.85\columnwidth}{l r r r r r}
    \toprule
    \multirow{2}{*}{Task}  & \multicolumn{3}{c}{Training targets}  & \multirow{2}{*}{\emph{Gold}} \\
    \cmidrule{2-4}
           & \multicolumn{1}{c}{T}  &  \multicolumn{1}{c}{MI} & TMI       &  \\
    \midrule
 CoLA          &  -0.4      &  4.0 & \textbf{10.3}      & \emph{14.3} \\
 SST-2         &  50.3      & \textbf{77.8} & 76.9      & \emph{94.0} \\
 MRPC          &  32.8      & 57.4 & \textbf{64.0}      & \emph{86.3} \\
 QQP           &  64.3      & \textbf{77.3} & 76.4      & \emph{91.2} \\
 MNLI-m        &  41.0      & 52.3 & \textbf{52.9}      & \emph{62.4} \\
 MNLI-mm       &  40.8      & 54.2 & \textbf{54.8}      & \emph{62.6} \\
 QNLI          &  50.1      & \textbf{61.0} & 59.9      & \emph{64.3} \\
 RTE           &  50.9      & \textbf{59.2} & 57.0      & \emph{70.0} \\
 WNLI          &  45.1      & 46.5 & \textbf{50.7}      & \emph{56.3} \\
 Avg. w/o WNLI &  41.7      & 54.4 & \textbf{55.9}      & \emph{66.8} \\
    \bottomrule
    \end{tabular*}
    }
    \caption{
    	Zero-shot evaluation of spoken GLUE tasks using the SpeechGLUE dataset.
	All results are accuracy scores, except for CoLA that uses MCC.
	The STS-B task is excluded because the LLM failed to provide interpretable results.
	}
    \label{tab:results_glue}
\end{table}

\begin{table}[]
    \center
    {
    \footnotesize
    \begin{tabular*}{0.85\columnwidth}{l r r r r r}
    \toprule
    \multirow{2}{*}{Languages}  & \multicolumn{3}{c}{Training targets}  & \multirow{2}{*}{\emph{Gold}} \\
    \cmidrule{2-4}
           & \multicolumn{1}{c}{T}  &  \multicolumn{1}{c}{MI} & TMI       &  \\
    \midrule
    BLOOM (9)     & 36.3 & \textbf{42.8} & \textbf{42.8} & \emph{54.2} \\
    Non-BLOOM (6) & 35.1 & \textbf{39.7} & 39.4 & \emph{43.9} \\
    All (15)      & 35.8 & \textbf{41.6} & 41.4 & \emph{50.1} \\
    \bottomrule
    \end{tabular*}
    }
    \caption{
    	Zero-shot evaluation of multilingual spoken NLI using the SpeechXNLI dataset.
	All results are accuracy scores.
    }
    \label{tab:results_xnli}
\end{table}

Tables \ref{tab:results_glue} and \ref{tab:results_xnli} provide the results of zero-shot
evaluation of BLOOMZMMS models on spoken GLUE tasks
in English using the SpeechGLUE dataset and on spoken NLI tasks in multiple languages
using the SpeechXNLI dataset.
It is worth noting that the combined TMI training targets result in better performance
on the English GLUE tasks, but have a mixed impact on the NLI tasks based on the
languages trained in BLOOM and those that were not.
For the BLOOM languages,
the TMI model equals the MI-only model in accuracy,
whereas it performs worse on the non-BLOOM languages.
Together with the SLT results, this observation again hints at
the effect of the LLM's weaknesses amplification during the transfer
from the text to speech modality.

\subsection{Visual Analysis}

\begin{figure*}[!ht]
	\begin{subfigure}[t]{.333\linewidth}
		\centerline{\includegraphics[scale=0.3]{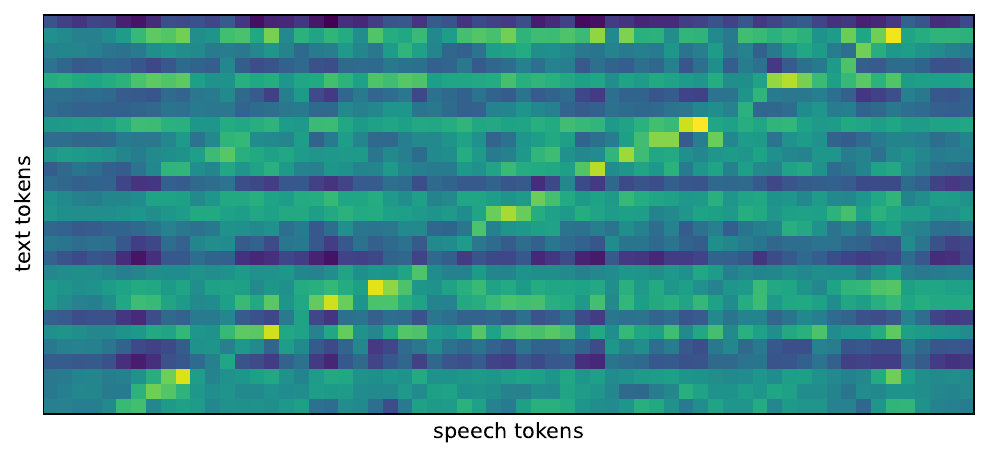}}
	\end{subfigure}%
	\begin{subfigure}[t]{.333\linewidth}
		\centerline{\includegraphics[scale=0.3]{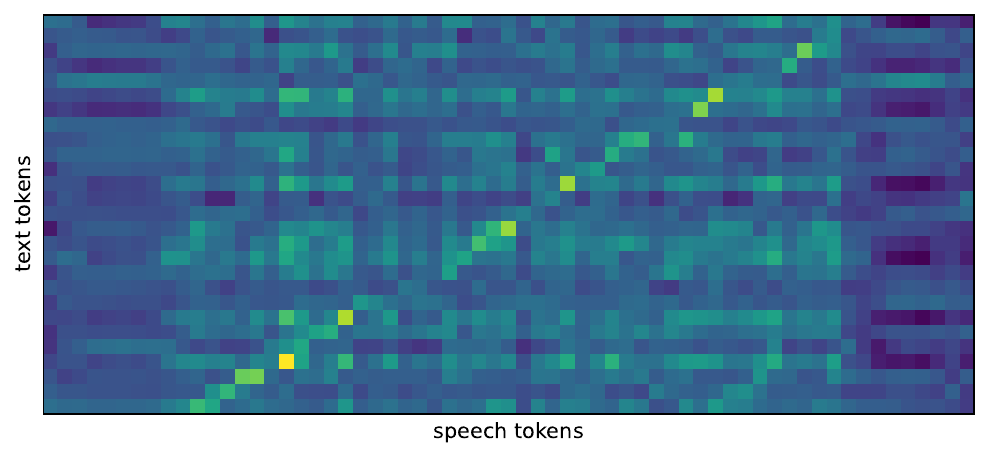}}
	\end{subfigure}%
	\begin{subfigure}[t]{.333\linewidth}
		\centerline{\includegraphics[scale=0.3]{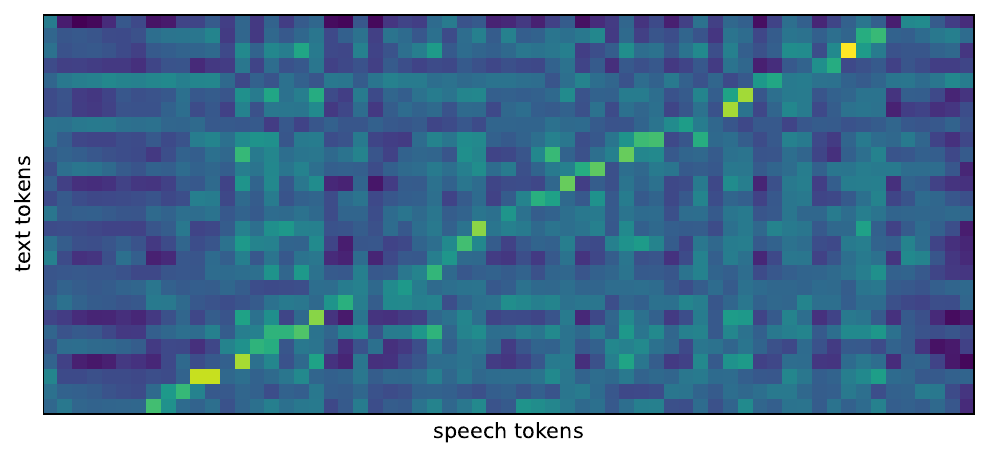}}
	\end{subfigure}
	\begin{subfigure}[t]{.333\linewidth}
		\centerline{\includegraphics[scale=0.3]{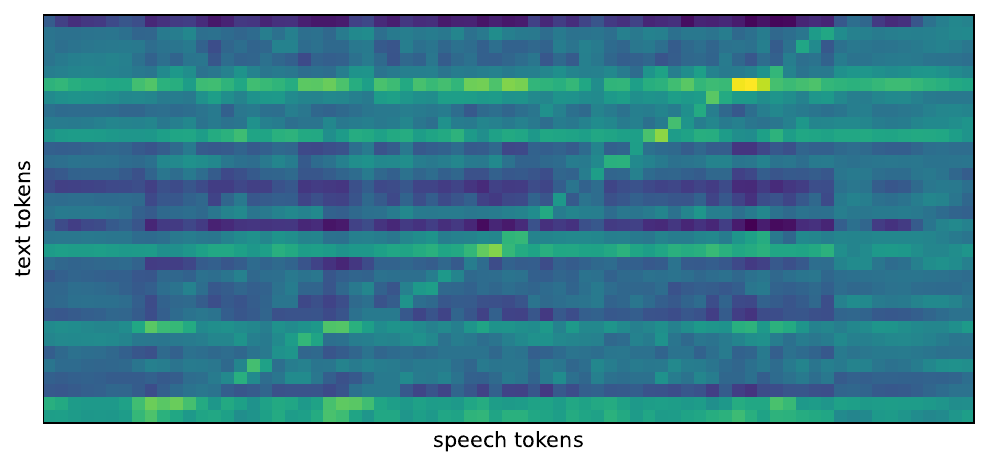}}
	\end{subfigure}%
	\begin{subfigure}[t]{.333\linewidth}
		\centerline{\includegraphics[scale=0.3]{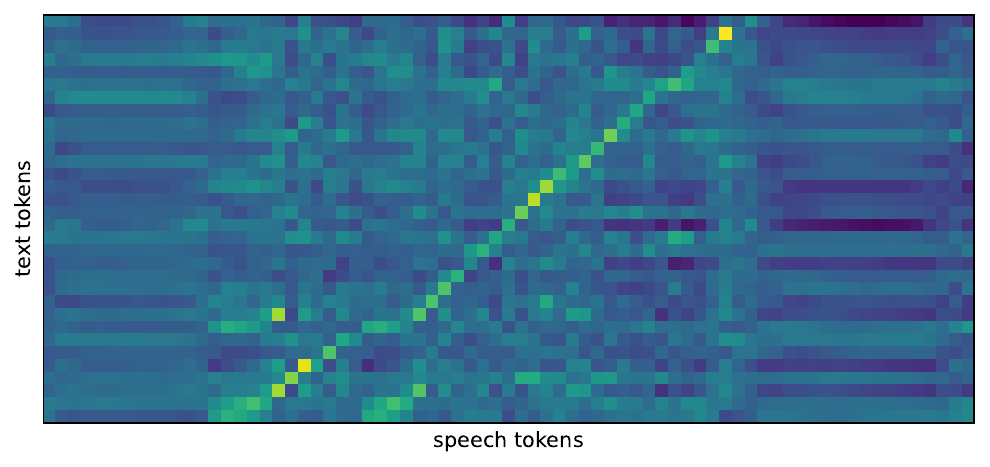}}
	\end{subfigure}%
	\begin{subfigure}[t]{.333\linewidth}
		\centerline{\includegraphics[scale=0.3]{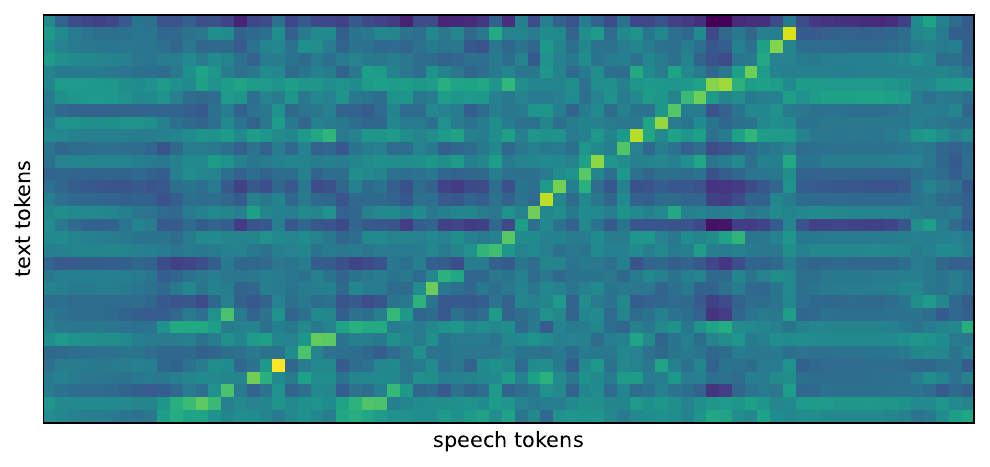}}
	\end{subfigure}
	\caption{
	Cosine similarity between the text and speech embeddings for two FLEURS evaluation utterances.
	Rows correspond to French and Finnish languages (seen and unseen by BLOOM).
	Columns represent the T, MI and TMI models.
	}
	\label{fig:embeddings}
\end{figure*}

Following the example of \cite{fathullah2023prompting}, we display
the cosine similarity between the text and speech embeddings
for the three variants of BLOOMZMMS for a French and a Finnish utterance
from the FLEURS evaluation dataset (Figure \ref{fig:embeddings}).
Consistent with the objective metrics from our experiments,
the model trained on the transcription targets shows the noisiest alignments
for the both languages, while the MI training targets offer better alignment
for a language unseen by BLOOM and the combined training targets
work better for a language seen by BLOOM.

\section{Conclusion}

In this paper we present BLOOMZMMS, a multilingual multitask speech processing
model that combines a multilingual LLM and a pretrained multilingual speech encoder.
Our investigation into two training strategies revealed their combined efficacy
in a broad spectrum of spoken language processing tasks,
a conclusion bolstered by zero-shot evaluations on multiple benchmarks.

\section*{Limitations}

Our setup is based on pretrained models and,
given that our experiments solely rely on ASR data
for supervision and the pretrained models remain frozen,
the performance in tasks beyond ASR
is limited by the capabilities of the utilized pretrained models.
For example, the SLT results
cannot be better than the translation results of the BLOOMZ model
on text input.

While we demonstrate the benefits of multi-instructional training in terms
of task generalization in transferring LLM abilities
from text to speech modality,
our evaluation is limited to a fixed collection of instructions.
It does not investigate the impact
of varying combinations of instructions more broadly
and whether the performance on a certain task depends on its presence
in the multi-instructional training data.
Furthermore, we do not compare synthetic label generation with
the use of ground truth labels,
a comparison that holds particular significance
for the SLT task.
A substantial amount of ground truth labeled data is
available for the SLT task.
Utilizing this data could likely enhance the model's
performance for this task, and potentially others as well.
Finally, the slight performance degradation observed
in the in-domain ASR dataset with TMI training
could potentially be mitigated by more effectively balancing
between transcription and multi-instructional data.

Our study is based on a small set of speech processing tasks,
and does not consider such tasks as spoken question answering,
spoken document summarization and other generative tasks.
In addition to that, our evaluation is restricted to the properties
of the used evaluation data. For the ASR and SLT
tasks, it is read speech recorded on a close distance microphone.
For the speech understanding tasks, we rely on a single speaker speech synthesis.
It should not be assumed that the proposed
model would work equally well or poorly for unseen tasks or
new recording conditions, such as far field noisy conversational
speech with possibly overlapping speakers.
Assuming the model's performance without empirical
testing in various scenarios could lead to risks,
particularly depending on its application.
This risk should be mitigated through
preliminary testing specific to each use case.
Additionally, it is advisable to cross-check
the model's outputs with independent information sources.

\bibliography{anthology,anthology_p2,refs}

\clearpage

\appendix
\onecolumn

\section{Evaluation Scores}

\subsection{Speech Recognition}
\begin{center}
\footnotesize
\topcaption{}
\bottomcaption{}
\tablecaption{Speech recognition results on the FLEURS evaluation dataset, broken down by language and training targets.}
\tablefirsthead{\toprule
\multirow{2}{*}{Language}&\multicolumn{3}{c}{WER, \%}&\multicolumn{3}{c}{CER, \%}\\
\cmidrule(r){2-4}\cmidrule{5-7}
&
\multicolumn{1}{c}{T}&\multicolumn{1}{c}{MI}&\multicolumn{1}{c}{TMI}&
\multicolumn{1}{c}{T}&\multicolumn{1}{c}{MI}&\multicolumn{1}{c}{TMI}\\
\midrule}
\tablehead{\toprule
\multirow{2}{*}{Language}&\multicolumn{3}{c}{WER, \%}&\multicolumn{3}{c}{CER, \%}\\
\cmidrule(r){2-4}\cmidrule{5-7}
&
\multicolumn{1}{c}{T}&\multicolumn{1}{c}{MI}&\multicolumn{1}{c}{TMI}&
\multicolumn{1}{c}{T}&\multicolumn{1}{c}{MI}&\multicolumn{1}{c}{TMI}\\
\midrule}
\tabletail{\midrule}
\tablelasttail{\\\bottomrule}
\begin{xtabular}{lrrrrrr}
Afrikaans & 24.55 & 73.55 & 25.11 & 11.48 & 49.00 & 11.60 \\
Amharic & 33.27 & 252.00 & 35.29 & 13.00 & 239.76 & 13.94 \\
Arabic & 28.21 & 72.32 & 26.87 & 12.50 & 55.77 & 11.46 \\
Armenian & 23.49 & 221.59 & 25.98 & 8.35 & 171.45 & 9.13 \\
Assamese & 20.67 & 111.94 & 22.43 & 16.12 & 97.06 & 17.59 \\
Asturian & 22.00 & 62.90 & 24.30 & 8.24 & 31.56 & 10.45 \\
Azerbaijani & 26.11 & 153.81 & 27.31 & 7.05 & 116.91 & 7.66 \\
Belarusian & 16.80 & 163.77 & 17.52 & 7.04 & 114.60 & 7.00 \\
Bengali & 5.36 & 58.87 & 6.02 & 4.12 & 54.77 & 4.82 \\
Bosnian & 15.62 & 68.29 & 15.97 & 5.57 & 44.96 & 5.73 \\
Bulgarian & 16.17 & 119.36 & 16.00 & 5.74 & 99.81 & 5.79 \\
Burmese & 46.24 & 92.22 & 47.42 & 39.87 & 76.31 & 40.82 \\
Cantonese Chinese & 42.94 & 142.09 & 37.88 & 19.18 & 92.69 & 17.77 \\
Catalan & 5.27 & 17.66 & 6.38 & 3.39 & 11.70 & 4.29 \\
Cebuano & 14.94 & 73.46 & 15.23 & 6.34 & 53.49 & 6.22 \\
Croatian & 22.09 & 81.07 & 21.93 & 11.64 & 53.73 & 11.66 \\
Czech & 14.08 & 85.25 & 14.46 & 5.19 & 55.41 & 5.22 \\
Danish & 25.80 & 85.27 & 25.48 & 10.53 & 53.11 & 9.78 \\
Dutch & 12.94 & 78.68 & 13.77 & 5.20 & 51.60 & 5.66 \\
English & 6.11 & 12.01 & 5.85 & 4.29 & 8.29 & 4.24 \\
Estonian & 15.03 & 87.18 & 16.26 & 4.13 & 53.03 & 7.21 \\
Filipino & 13.05 & 63.70 & 13.64 & 5.13 & 45.66 & 5.25 \\
Finnish & 16.65 & 90.32 & 17.75 & 4.58 & 54.73 & 5.00 \\
French & 5.20 & 13.61 & 5.04 & 3.29 & 9.25 & 3.36 \\
Fula & 51.09 & 106.50 & 50.40 & 20.47 & 77.03 & 18.87 \\
Galician & 15.30 & 62.75 & 15.18 & 6.88 & 29.57 & 6.08 \\
Ganda & 41.03 & 127.78 & 41.32 & 10.67 & 85.24 & 10.78 \\
Georgian & 30.98 & 260.91 & 31.80 & 10.44 & 160.02 & 10.76 \\
German & 11.66 & 81.32 & 11.45 & 4.93 & 52.19 & 4.76 \\
Greek & 20.99 & 145.51 & 22.68 & 8.55 & 123.39 & 9.50 \\
Gujarati & 14.01 & 94.25 & 11.62 & 10.55 & 88.46 & 8.97 \\
Hausa & 25.50 & 92.04 & 25.54 & 9.14 & 65.99 & 8.87 \\
Hebrew & 53.53 & 196.05 & 53.93 & 24.07 & 152.62 & 25.17 \\
Hindi & 10.74 & 42.76 & 9.11 & 8.46 & 39.51 & 7.17 \\
Hungarian & 21.07 & 106.88 & 21.51 & 6.82 & 66.61 & 6.96 \\
Icelandic & 35.32 & 115.87 & 36.01 & 10.52 & 67.09 & 11.18 \\
Igbo & 41.68 & 134.93 & 42.42 & 22.65 & 112.95 & 23.82 \\
Indonesian & 5.84 & 30.87 & 5.37 & 3.84 & 22.66 & 3.77 \\
Irish & 58.24 & 122.34 & 60.19 & 28.42 & 82.76 & 29.45 \\
Italian & 7.16 & 56.15 & 7.25 & 3.77 & 34.05 & 3.78 \\
Japanese & 94.35 & 321.61 & 101.43 & 23.82 & 155.45 & 28.23 \\
Javanese & 19.71 & 132.97 & 21.40 & 7.25 & 99.08 & 8.57 \\
Kabuverdianu & 20.70 & 79.69 & 19.64 & 8.01 & 52.42 & 7.16 \\
Kamba & 45.84 & 147.72 & 47.05 & 17.85 & 107.02 & 20.67 \\
Kannada & 24.16 & 100.43 & 15.65 & 18.69 & 96.71 & 12.84 \\
Kazakh & 17.26 & 155.52 & 17.55 & 5.78 & 117.24 & 5.84 \\
Khmer & 50.99 & 107.49 & 59.89 & 29.21 & 97.01 & 35.90 \\
Korean & 42.39 & 164.91 & 48.85 & 16.70 & 183.41 & 20.96 \\
Kyrgyz & 18.16 & 164.98 & 18.33 & 5.25 & 124.54 & 5.08 \\
Lao & 69.97 & 120.78 & 74.54 & 50.98 & 105.17 & 54.48 \\
Latvian & 15.73 & 93.54 & 16.50 & 4.94 & 55.89 & 5.27 \\
Lingala & 14.15 & 116.21 & 15.66 & 6.88 & 91.66 & 8.86 \\
Lithuanian & 20.96 & 102.37 & 21.59 & 6.64 & 63.12 & 6.63 \\
Luo & 26.80 & 81.87 & 26.44 & 6.77 & 56.46 & 7.27 \\
Luxembourgish & 34.45 & 140.89 & 36.66 & 11.87 & 90.38 & 13.19 \\
Macedonian & 11.29 & 124.75 & 11.04 & 4.17 & 101.96 & 4.07 \\
Malay & 16.48 & 77.65 & 17.55 & 8.86 & 57.36 & 8.94 \\
Malayalam & 17.47 & 95.27 & 14.61 & 14.11 & 89.51 & 11.89 \\
Maltese & 16.37 & 104.21 & 16.38 & 6.33 & 75.53 & 5.77 \\
Mandarin Chinese & 36.12 & 103.75 & 32.73 & 15.63 & 58.45 & 14.24 \\
Maori & 22.50 & 94.78 & 22.79 & 9.90 & 69.21 & 9.65 \\
Marathi & 11.13 & 83.43 & 9.66 & 8.76 & 72.61 & 7.55 \\
Mongolian & 33.30 & 159.58 & 34.57 & 10.37 & 135.18 & 11.24 \\
Nepali & 13.32 & 77.91 & 9.77 & 10.28 & 68.06 & 7.48 \\
Northern-Sotho & 27.13 & 99.29 & 27.03 & 14.08 & 74.31 & 13.88 \\
Norwegian & 19.26 & 67.88 & 20.11 & 6.95 & 42.36 & 7.05 \\
Nyanja & 35.21 & 116.14 & 34.56 & 13.08 & 82.44 & 12.56 \\
Occitan & 31.98 & 89.00 & 33.23 & 13.11 & 53.19 & 13.24 \\
Oriya & 26.79 & 113.60 & 24.98 & 19.84 & 100.74 & 18.93 \\
Oromo & 64.94 & 105.36 & 68.18 & 17.51 & 59.86 & 18.12 \\
Pashto & 48.14 & 190.56 & 52.98 & 21.68 & 138.96 & 24.79 \\
Persian & 18.07 & 128.77 & 18.63 & 6.92 & 94.73 & 6.99 \\
Polish & 13.82 & 104.07 & 15.22 & 5.56 & 71.10 & 5.90 \\
Portuguese & 4.45 & 17.50 & 4.71 & 3.09 & 12.42 & 3.31 \\
Punjabi & 21.03 & 111.35 & 20.46 & 15.41 & 97.24 & 15.56 \\
Romanian & 14.51 & 88.13 & 15.67 & 6.08 & 54.48 & 6.40 \\
Russian & 19.32 & 123.15 & 19.16 & 6.52 & 96.07 & 6.10 \\
Serbian & 57.70 & 131.08 & 55.88 & 47.03 & 109.11 & 45.26 \\
Shona & 22.28 & 141.53 & 24.29 & 7.35 & 91.81 & 9.15 \\
Sindhi & 28.99 & 181.61 & 31.50 & 12.44 & 145.41 & 14.03 \\
Slovak & 12.17 & 85.16 & 12.52 & 4.87 & 53.21 & 4.99 \\
Slovenian & 18.38 & 86.15 & 18.37 & 6.91 & 59.06 & 6.61 \\
Somali & 45.93 & 122.01 & 45.82 & 16.41 & 76.23 & 17.14 \\
Sorani-Kurdish & 39.09 & 139.68 & 40.47 & 11.56 & 107.75 & 12.02 \\
Spanish & 3.65 & 10.52 & 3.66 & 2.53 & 7.74 & 2.62 \\
Swahili & 10.81 & 85.55 & 12.31 & 5.84 & 65.39 & 7.02 \\
Swedish & 21.50 & 78.59 & 21.92 & 7.44 & 49.06 & 7.52 \\
Tajik & 17.81 & 166.92 & 18.54 & 6.87 & 125.87 & 7.16 \\
Tamil & 14.14 & 77.87 & 9.80 & 11.92 & 73.81 & 7.78 \\
Telugu & 22.68 & 99.79 & 19.13 & 17.32 & 94.83 & 15.12 \\
Thai & 36.60 & 161.15 & 38.97 & 15.76 & 100.18 & 17.44 \\
Turkish & 18.39 & 129.47 & 19.59 & 5.30 & 97.46 & 5.82 \\
Ukrainian & 17.86 & 134.65 & 18.04 & 5.12 & 105.12 & 4.86 \\
Umbundu & 46.97 & 155.61 & 47.71 & 16.44 & 104.41 & 17.33 \\
Urdu & 96.48 & 129.51 & 86.06 & 49.48 & 86.81 & 44.44 \\
Uzbek & 26.77 & 99.35 & 26.61 & 8.58 & 65.17 & 8.34 \\
Vietnamese & 25.37 & 65.35 & 23.84 & 20.20 & 55.97 & 19.19 \\
Welsh & 28.34 & 75.28 & 29.78 & 10.49 & 44.37 & 10.80 \\
Wolof & 35.70 & 104.97 & 37.68 & 14.97 & 78.91 & 17.38 \\
Xhosa & 34.67 & 162.98 & 38.63 & 10.90 & 99.80 & 14.64 \\
Yoruba & 55.10 & 129.76 & 65.13 & 29.54 & 109.40 & 39.49 \\
Zulu & 31.56 & 148.55 & 33.56 & 10.46 & 93.27 & 13.11 \\
\midrule
Median & 21.75 & 104.14 & 21.93 & 9.00 & 76.27 & 9.05 \\
Average & 26.70 & 110.47 & 27.20 & 12.03 & 80.94 & 12.41
\end{xtabular}

\end{center}

\begin{center}
\footnotesize
\topcaption{}
\bottomcaption{}
\tablecaption{Speech recognition results on the Multilingual LibriSpeech evaluation dataset, broken down by language and training targets.}
\tablefirsthead{\toprule
\multirow{2}{*}{Language}&\multicolumn{3}{c}{WER, \%}&\multicolumn{3}{c}{CER, \%}\\
\cmidrule(r){2-4}\cmidrule{5-7}
&
\multicolumn{1}{c}{T}&\multicolumn{1}{c}{MI}&\multicolumn{1}{c}{TMI}&
\multicolumn{1}{c}{T}&\multicolumn{1}{c}{MI}&\multicolumn{1}{c}{TMI}\\
\midrule}
\tablehead{\toprule
\multirow{2}{*}{Language}&\multicolumn{3}{c}{WER, \%}&\multicolumn{3}{c}{CER, \%}\\
\cmidrule(r){2-4}\cmidrule{5-7}
&
\multicolumn{1}{c}{T}&\multicolumn{1}{c}{MI}&\multicolumn{1}{c}{TMI}&
\multicolumn{1}{c}{T}&\multicolumn{1}{c}{MI}&\multicolumn{1}{c}{TMI}\\
\midrule}
\tabletail{\midrule}
\tablelasttail{\\\bottomrule}
\begin{xtabular}{lrrrrrr}
Dutch & 13.03 & 71.04 & 14.10 & 4.15 & 51.80 & 4.99 \\
English & 36.14 & 23.79 & 26.08 & 26.51 & 17.76 & 18.77 \\
French & 21.92 & 23.41 & 17.28 & 15.71 & 17.61 & 12.35 \\
German & 10.62 & 76.19 & 11.29 & 4.28 & 52.03 & 4.64 \\
Italian & 13.34 & 48.41 & 14.18 & 3.72 & 29.77 & 4.26 \\
Polish & 8.89 & 93.88 & 9.25 & 2.41 & 63.78 & 2.29 \\
Portuguese & 33.68 & 28.03 & 25.53 & 23.07 & 20.32 & 17.10 \\
Spanish & 16.47 & 24.67 & 13.57 & 10.64 & 19.15 & 9.10 \\
\midrule
Median & 14.91 & 38.22 & 14.14 & 7.46 & 25.04 & 7.04 \\
Average & 19.26 & 48.68 & 16.41 & 11.31 & 34.03 & 9.18
\end{xtabular}

\end{center}

\begin{center}
\footnotesize
\topcaption{}
\bottomcaption{}
\tablecaption{Speech recognition results on the VoxPopuli evaluation dataset, broken down by language and training targets.}
\tablefirsthead{\toprule
\multirow{2}{*}{Language}&\multicolumn{3}{c}{WER, \%}&\multicolumn{3}{c}{CER, \%}\\
\cmidrule(r){2-4}\cmidrule{5-7}
&
\multicolumn{1}{c}{T}&\multicolumn{1}{c}{MI}&\multicolumn{1}{c}{TMI}&
\multicolumn{1}{c}{T}&\multicolumn{1}{c}{MI}&\multicolumn{1}{c}{TMI}\\
\midrule}
\tablehead{\toprule
\multirow{2}{*}{Language}&\multicolumn{3}{c}{WER, \%}&\multicolumn{3}{c}{CER, \%}\\
\cmidrule(r){2-4}\cmidrule{5-7}
&
\multicolumn{1}{c}{T}&\multicolumn{1}{c}{MI}&\multicolumn{1}{c}{TMI}&
\multicolumn{1}{c}{T}&\multicolumn{1}{c}{MI}&\multicolumn{1}{c}{TMI}\\
\midrule}
\tabletail{\midrule}
\tablelasttail{\\\bottomrule}
\begin{xtabular}{lrrrrrr}
Croatian & 25.24 & 92.13 & 23.74 & 10.55 & 65.04 & 10.94 \\
Czech & 14.25 & 108.21 & 16.18 & 7.70 & 70.56 & 9.40 \\
Dutch & 24.72 & 98.76 & 21.62 & 16.34 & 70.28 & 13.77 \\
English & 21.00 & 18.31 & 16.14 & 15.53 & 13.79 & 11.62 \\
Estonian & 17.73 & 133.44 & 17.73 & 7.75 & 89.95 & 6.71 \\
Finnish & 21.25 & 115.89 & 20.80 & 10.35 & 71.74 & 9.22 \\
French & 23.12 & 25.54 & 18.91 & 16.60 & 19.93 & 13.79 \\
German & 25.78 & 101.91 & 24.96 & 17.28 & 68.95 & 16.74 \\
Hungarian & 18.86 & 119.21 & 19.02 & 8.70 & 77.14 & 8.27 \\
Italian & 26.17 & 78.50 & 28.29 & 19.64 & 55.93 & 20.43 \\
Latvian & 25.61 & 145.96 & 31.86 & 13.73 & 102.17 & 22.62 \\
Polish & 17.66 & 123.45 & 17.00 & 11.32 & 86.39 & 11.34 \\
Romanian & 18.61 & 99.78 & 20.88 & 8.59 & 62.19 & 9.72 \\
Slovak & 17.79 & 110.26 & 18.21 & 9.80 & 69.64 & 9.51 \\
Slovenian & 33.59 & 102.96 & 31.15 & 25.08 & 80.19 & 24.66 \\
Spanish & 19.79 & 22.07 & 16.76 & 14.43 & 16.40 & 11.82 \\
\midrule
Median & 21.12 & 102.43 & 19.91 & 12.52 & 69.96 & 11.48 \\
Average & 21.95 & 93.52 & 21.45 & 13.34 & 63.77 & 13.16
\end{xtabular}

\end{center}

\newpage
\subsection{Speech Translation}
\begin{center}
\footnotesize
\topcaption{}
\bottomcaption{}
\tablecaption{Speech translation results on the CoVoST-2 English $\rightarrow$ X evaluation dataset, broken down by target language and training targets.}
\tablefirsthead{\toprule
\multirow{2}{*}{Language}&\multicolumn{3}{c}{BLEU}&\multicolumn{3}{c}{chrF}\\
\cmidrule(r){2-4}\cmidrule{5-7}
&
\multicolumn{1}{c}{T}&\multicolumn{1}{c}{MI}&\multicolumn{1}{c}{TMI}&
\multicolumn{1}{c}{T}&\multicolumn{1}{c}{MI}&\multicolumn{1}{c}{TMI}\\
\midrule}
\tablehead{\toprule
\multirow{2}{*}{Language}&\multicolumn{3}{c}{BLEU}&\multicolumn{3}{c}{chrF}\\
\cmidrule(r){2-4}\cmidrule{5-7}
&
\multicolumn{1}{c}{T}&\multicolumn{1}{c}{MI}&\multicolumn{1}{c}{TMI}&
\multicolumn{1}{c}{T}&\multicolumn{1}{c}{MI}&\multicolumn{1}{c}{TMI}\\
\midrule}
\tabletail{\midrule}
\tablelasttail{\\\bottomrule}
\begin{xtabular}{lrrrrrr}
Arabic & 0.93 & 9.33 & 8.69 & 3.88 & 34.00 & 33.56 \\
Catalan & 2.27 & 20.63 & 20.97 & 21.42 & 46.17 & 46.35 \\
Estonian & 0.47 & 0.25 & 0.31 & 15.78 & 13.78 & 14.37 \\
German & 1.23 & 4.85 & 4.85 & 18.04 & 26.86 & 26.67 \\
Indonesian & 1.93 & 21.66 & 22.53 & 17.26 & 49.34 & 49.67 \\
Japanese & 0.02 & 0.00 & 0.03 & 0.93 & 4.52 & 3.60 \\
Latvian & 0.00 & 0.08 & 0.14 & 0.00 & 9.60 & 11.11 \\
Mandarin Chinese & 0.14 & 0.00 & 0.31 & 4.90 & 16.43 & 17.92 \\
Mongolian & 0.00 & 0.05 & 0.08 & 0.00 & 0.77 & 0.80 \\
Persian & 0.12 & 0.00 & 0.04 & 1.08 & 8.72 & 8.20 \\
Slovenian & 0.00 & 0.15 & 0.19 & 1.67 & 10.55 & 11.63 \\
Swedish & 0.00 & 2.85 & 3.22 & 1.00 & 19.54 & 20.62 \\
Tamil & 0.00 & 2.68 & 2.66 & 2.51 & 31.71 & 31.33 \\
Turkish & 0.00 & 0.23 & 0.23 & 1.60 & 10.26 & 11.76 \\
Welsh & 0.90 & 0.34 & 0.44 & 15.33 & 11.86 & 12.54 \\
\midrule
Median & 0.12 & 0.25 & 0.31 & 2.51 & 13.78 & 14.37 \\
Average & 0.53 & 4.21 & 4.31 & 7.03 & 19.61 & 20.01
\end{xtabular}

\end{center}

\begin{center}
\footnotesize
\topcaption{}
\bottomcaption{}
\tablecaption{Speech translation results on the CoVoST-2 X $\rightarrow$ English evaluation dataset, broken down by source language and training targets.}
\tablefirsthead{\toprule
& \multirow{2}{*}{Language}&\multicolumn{3}{c}{BLEU}&\multicolumn{3}{c}{chrF}\\
\cmidrule(r){3-5}\cmidrule{6-8}
&
& \multicolumn{1}{c}{T}&\multicolumn{1}{c}{MI}&\multicolumn{1}{c}{TMI}&
\multicolumn{1}{c}{T}&\multicolumn{1}{c}{MI}&\multicolumn{1}{c}{TMI}\\
\midrule}
\tablehead{\toprule
& \multirow{2}{*}{Language}&\multicolumn{3}{c}{BLEU}&\multicolumn{3}{c}{chrF}\\
\cmidrule(r){3-5}\cmidrule{6-8}
&
& \multicolumn{1}{c}{T}&\multicolumn{1}{c}{MI}&\multicolumn{1}{c}{TMI}&
\multicolumn{1}{c}{T}&\multicolumn{1}{c}{MI}&\multicolumn{1}{c}{TMI}\\
\midrule}
\tabletail{\midrule}
\tablelasttail{\\\bottomrule}
\begin{xtabular}{llrrrrrr}
\multirow{4}{*}{High} & French & 4.29 & 30.11 & 31.13 & 29.08 & 54.46 & 55.45 \\
& German & 1.99 & 18.92 & 19.27 & 21.70 & 41.70 & 41.84 \\
& Spanish & 4.66 & 33.64 & 34.78 & 28.68 & 58.59 & 59.39 \\
& Catalan & 2.17 & 27.66 & 28.12 & 23.41 & 52.06 & 52.43 \\
\midrule
\multirow{5}{*}{Mid} & Persian & 0.06 & 1.46 & 1.34 & 0.38 & 15.65 & 15.76 \\
& Italian & 1.92 & 26.91 & 27.30 & 26.27 & 52.28 & 52.35 \\
& Russian & 0.92 & 24.55 & 23.22 & 3.82 & 49.13 & 47.80 \\
& Portugese & 15.68 & 41.74 & 42.58 & 32.76 & 62.82 & 63.33 \\
& Mandarin Chinese & 0.00 & 10.21 & 10.87 & 0.03 & 30.12 & 30.74 \\
\midrule
\multirow{12}{*}{Low} & Turkish & 0.00 & 1.25 & 1.30 & 11.23 & 14.36 & 14.65 \\
& Arabic & 21.16 & 29.11 & 29.67 & 34.86 & 50.41 & 50.69 \\
& Estonian & 0.12 & 0.41 & 0.52 & 16.56 & 15.20 & 16.31 \\
& Mongolian & 0.00 & 0.17 & 0.00 & 0.68 & 13.12 & 13.10 \\
& Dutch & 1.03 & 15.35 & 15.29 & 20.37 & 34.78 & 34.40 \\
& Swedish & 0.56 & 8.75 & 10.32 & 13.36 & 24.29 & 25.09 \\
& Latvian & 0.00 & 0.83 & 0.79 & 9.30 & 11.77 & 11.08 \\
& Slovenian & 0.00 & 2.91 & 3.47 & 11.14 & 14.70 & 14.94 \\
& Tamil & 0.00 & 2.38 & 2.63 & 0.43 & 17.14 & 17.15 \\
& Japanese & 0.00 & 7.65 & 9.45 & 0.37 & 22.40 & 25.70 \\
& Indonesian & 8.31 & 32.53 & 34.36 & 20.85 & 49.26 & 51.67 \\
& Welsh & 0.57 & 0.49 & 0.87 & 13.46 & 12.41 & 13.66 \\
\midrule
& Median & 0.57 & 10.21 & 10.87 & 13.46 & 30.12 & 30.74 \\
& Average & 3.02 & 15.10 & 15.58 & 15.18 & 33.17 & 33.69
\end{xtabular}

\end{center}

\newpage
\begin{center}
\footnotesize
\topcaption{}
\bottomcaption{}
\tablecaption{Speech translation results on the FLEURS English $\rightarrow$ X evaluation dataset, broken down by target language and training targets.}
\tablefirsthead{\toprule
\multirow{2}{*}{Language}&\multicolumn{3}{c}{BLEU}&\multicolumn{3}{c}{chrF}\\
\cmidrule(r){2-4}\cmidrule{5-7}
&
\multicolumn{1}{c}{T}&\multicolumn{1}{c}{MI}&\multicolumn{1}{c}{TMI}&
\multicolumn{1}{c}{T}&\multicolumn{1}{c}{MI}&\multicolumn{1}{c}{TMI}\\
\midrule}
\tablehead{\toprule
\multirow{2}{*}{Language}&\multicolumn{3}{c}{BLEU}&\multicolumn{3}{c}{chrF}\\
\cmidrule(r){2-4}\cmidrule{5-7}
&
\multicolumn{1}{c}{T}&\multicolumn{1}{c}{MI}&\multicolumn{1}{c}{TMI}&
\multicolumn{1}{c}{T}&\multicolumn{1}{c}{MI}&\multicolumn{1}{c}{TMI}\\
\midrule}
\tabletail{\midrule}
\tablelasttail{\\\bottomrule}
\begin{xtabular}{lrrrrrr}
Amharic & 0.25 & 0.00 & 0.00 & 0.64 & 0.47 & 0.52 \\
Arabic & 6.27 & 13.09 & 13.72 & 27.62 & 42.08 & 42.79 \\
Armenian & 0.26 & 0.00 & 0.00 & 0.48 & 0.45 & 0.43 \\
Assamese & 9.46 & 16.45 & 16.73 & 14.95 & 44.10 & 41.78 \\
Asturian & 2.07 & 9.81 & 10.16 & 25.38 & 44.77 & 44.63 \\
Azerbaijani & 0.46 & 0.00 & 0.19 & 14.31 & 12.64 & 13.88 \\
Belarusian & 0.29 & 0.00 & 0.21 & 0.70 & 1.20 & 1.12 \\
Bengali & 37.46 & 48.13 & 51.21 & 45.43 & 69.01 & 71.23 \\
Bosnian & 1.13 & 0.67 & 0.65 & 19.49 & 17.84 & 18.37 \\
Bulgarian & 0.54 & 0.33 & 0.33 & 0.79 & 1.54 & 1.49 \\
Burmese & 0.00 & 0.00 & 0.00 & 0.55 & 0.48 & 0.49 \\
Cantonese Chinese & 0.43 & 4.45 & 5.22 & 14.51 & 38.15 & 38.02 \\
Catalan & 35.35 & 49.12 & 50.26 & 53.42 & 70.47 & 71.09 \\
Cebuano & 2.61 & 1.63 & 2.07 & 19.66 & 16.63 & 18.27 \\
Croatian & 1.01 & 0.41 & 0.47 & 19.30 & 15.23 & 16.57 \\
Czech & 1.26 & 0.60 & 0.67 & 18.27 & 14.39 & 14.96 \\
Danish & 2.03 & 1.58 & 1.82 & 24.57 & 21.05 & 23.28 \\
Dutch & 1.85 & 1.59 & 1.54 & 23.85 & 22.04 & 22.49 \\
Estonian & 1.00 & 0.70 & 0.61 & 19.36 & 17.82 & 18.23 \\
Filipino & 2.62 & 1.45 & 1.56 & 18.29 & 16.97 & 17.39 \\
Finnish & 0.51 & 0.16 & 0.27 & 17.34 & 13.67 & 15.35 \\
French & 48.16 & 47.03 & 46.79 & 64.13 & 68.64 & 68.29 \\
Fula & 1.66 & 0.82 & 1.10 & 19.18 & 16.60 & 17.85 \\
Galician & 3.67 & 7.35 & 7.70 & 28.64 & 42.04 & 41.92 \\
Ganda & 2.22 & 1.29 & 1.39 & 17.06 & 15.40 & 16.04 \\
Georgian & 0.45 & 0.17 & 0.19 & 0.81 & 0.68 & 0.75 \\
German & 1.78 & 5.21 & 4.74 & 21.93 & 29.53 & 29.10 \\
Greek & 0.61 & 0.35 & 0.42 & 1.02 & 6.36 & 4.97 \\
Gujarati & 37.03 & 44.08 & 46.30 & 42.58 & 63.97 & 65.91 \\
Hausa & 1.55 & 0.80 & 0.77 & 16.56 & 14.06 & 14.79 \\
Hebrew & 0.57 & 0.23 & 0.36 & 1.09 & 1.41 & 1.26 \\
Hindi & 44.51 & 41.86 & 42.28 & 49.76 & 62.36 & 62.57 \\
Hungarian & 0.79 & 0.40 & 0.44 & 16.58 & 14.26 & 15.08 \\
Icelandic & 0.67 & 0.39 & 0.46 & 16.05 & 14.39 & 15.17 \\
Igbo & 2.06 & 2.66 & 2.88 & 17.11 & 18.26 & 18.39 \\
Indonesian & 39.06 & 50.40 & 50.81 & 54.08 & 71.86 & 71.92 \\
Irish & 1.65 & 1.06 & 1.12 & 17.19 & 15.00 & 15.54 \\
Italian & 2.07 & 8.61 & 8.02 & 25.28 & 36.03 & 35.54 \\
Japanese & 0.00 & 0.00 & 0.00 & 0.98 & 4.60 & 3.71 \\
Javanese & 2.05 & 2.87 & 2.74 & 20.80 & 28.13 & 28.13 \\
Kabuverdianu & 1.60 & 1.15 & 1.15 & 21.76 & 21.05 & 20.78 \\
Kamba & 2.47 & 1.24 & 1.34 & 18.30 & 12.48 & 14.10 \\
Kannada & 13.18 & 34.07 & 33.41 & 21.44 & 58.91 & 58.71 \\
Kazakh & 0.31 & 0.00 & 0.19 & 0.79 & 0.73 & 0.77 \\
Khmer & 0.78 & 0.49 & 0.56 & 2.59 & 2.22 & 2.29 \\
Korean & 0.51 & 0.17 & 0.33 & 2.61 & 1.75 & 2.39 \\
Kyrgyz & 0.21 & 0.00 & 0.00 & 0.78 & 0.70 & 0.72 \\
Lao & 1.37 & 0.93 & 1.03 & 3.65 & 3.04 & 3.21 \\
Latvian & 0.51 & 0.29 & 0.30 & 16.61 & 14.39 & 15.32 \\
Lingala & 1.93 & 4.20 & 3.70 & 17.99 & 23.79 & 22.38 \\
Lithuanian & 0.73 & 0.46 & 0.50 & 17.79 & 14.21 & 14.94 \\
Luo & 1.85 & 1.16 & 1.04 & 19.11 & 17.04 & 17.75 \\
Luxembourgish & 1.37 & 0.76 & 0.85 & 22.00 & 21.29 & 21.39 \\
Macedonian & 0.44 & 0.29 & 0.28 & 0.79 & 0.69 & 0.84 \\
Malay & 2.78 & 8.70 & 8.39 & 21.47 & 33.24 & 32.56 \\
Malayalam & 17.57 & 37.10 & 35.49 & 28.47 & 63.61 & 62.40 \\
Maltese & 1.68 & 1.23 & 1.15 & 20.40 & 18.72 & 19.21 \\
Mandarin Chinese & 2.82 & 3.27 & 5.90 & 38.25 & 32.76 & 36.63 \\
Maori & 1.85 & 1.31 & 1.52 & 17.99 & 17.35 & 17.56 \\
Marathi & 32.62 & 35.88 & 34.27 & 40.74 & 59.79 & 59.00 \\
Mongolian & 0.36 & 0.24 & 0.00 & 0.63 & 0.55 & 0.56 \\
Nepali & 19.35 & 40.99 & 42.16 & 26.63 & 64.08 & 63.41 \\
Northern-Sotho & 2.55 & 3.66 & 3.11 & 18.83 & 19.75 & 18.79 \\
Norwegian & 1.37 & 1.46 & 1.54 & 23.22 & 22.77 & 23.29 \\
Nyanja & 2.85 & 2.07 & 2.02 & 19.25 & 20.81 & 20.07 \\
Occitan & 1.67 & 3.73 & 3.51 & 25.67 & 34.52 & 33.53 \\
Oriya & 0.47 & 0.13 & 0.28 & 0.57 & 0.43 & 0.51 \\
Oromo & 0.00 & 0.00 & 0.00 & 14.18 & 13.03 & 13.84 \\
Pashto & 0.00 & 0.00 & 0.00 & 1.14 & 1.49 & 1.54 \\
Persian & 0.00 & 0.00 & 0.27 & 2.65 & 11.47 & 11.90 \\
Polish & 0.90 & 0.47 & 0.57 & 17.06 & 14.28 & 15.81 \\
Portuguese & 54.85 & 50.25 & 51.00 & 69.09 & 70.93 & 71.39 \\
Punjabi & 20.37 & 42.40 & 42.00 & 22.38 & 59.49 & 59.60 \\
Romanian & 1.40 & 1.30 & 1.40 & 23.50 & 21.84 & 22.65 \\
Russian & 0.51 & 0.97 & 0.99 & 0.72 & 6.75 & 8.30 \\
Serbian & 0.41 & 0.25 & 0.24 & 0.77 & 0.61 & 0.67 \\
Shona & 2.01 & 1.22 & 1.40 & 18.18 & 17.39 & 17.94 \\
Sindhi & 0.40 & 0.24 & 0.19 & 0.71 & 1.62 & 2.87 \\
Slovak & 1.08 & 0.54 & 0.75 & 18.24 & 14.86 & 15.72 \\
Slovenian & 1.05 & 0.44 & 0.50 & 18.90 & 14.85 & 16.05 \\
Somali & 1.82 & 1.17 & 1.34 & 15.50 & 14.17 & 14.73 \\
Sorani-Kurdish & 0.20 & 0.00 & 0.00 & 0.42 & 0.37 & 0.40 \\
Spanish & 30.80 & 29.44 & 28.79 & 51.37 & 56.27 & 55.64 \\
Swahili & 14.55 & 25.42 & 24.09 & 31.47 & 53.62 & 51.67 \\
Swedish & 1.65 & 1.68 & 1.67 & 24.06 & 22.28 & 23.66 \\
Tajik & 0.24 & 0.19 & 0.24 & 0.82 & 0.73 & 0.76 \\
Tamil & 26.38 & 52.30 & 49.51 & 34.49 & 73.57 & 72.05 \\
Telugu & 22.59 & 43.72 & 42.08 & 30.87 & 65.49 & 65.33 \\
Thai & 0.25 & 0.17 & 0.20 & 1.22 & 1.04 & 1.08 \\
Turkish & 0.78 & 0.47 & 0.54 & 17.27 & 15.20 & 16.30 \\
Ukrainian & 0.40 & 0.19 & 0.20 & 0.68 & 0.75 & 0.94 \\
Umbundu & 0.92 & 0.37 & 0.51 & 15.53 & 10.79 & 12.96 \\
Urdu & 26.11 & 35.76 & 37.54 & 31.93 & 55.39 & 55.79 \\
Uzbek & 0.33 & 0.25 & 0.32 & 16.21 & 15.02 & 16.05 \\
Vietnamese & 36.77 & 46.80 & 48.49 & 45.92 & 61.95 & 63.22 \\
Welsh & 1.50 & 0.94 & 1.14 & 18.45 & 16.36 & 17.32 \\
Wolof & 1.25 & 1.08 & 1.15 & 18.73 & 17.96 & 18.39 \\
Xhosa & 1.62 & 1.13 & 1.12 & 18.85 & 17.90 & 18.39 \\
Yoruba & 1.83 & 3.62 & 3.52 & 14.02 & 18.54 & 18.18 \\
Zulu & 1.13 & 0.98 & 0.77 & 17.29 & 17.21 & 17.22 \\
\midrule
Median & 1.53 & 1.11 & 1.13 & 18.25 & 17.00 & 17.66 \\
Average & 6.67 & 8.95 & 9.03 & 18.71 & 23.69 & 24.01
\end{xtabular}

\end{center}

\newpage
\begin{center}
\footnotesize
\topcaption{}
\bottomcaption{}
\tablecaption{Speech translation results on FLEURS X $\rightarrow$ English evaluation dataset, broken down by source language and training targets.}
\tablefirsthead{\toprule
\multirow{2}{*}{Language}&\multicolumn{3}{c}{BLEU}&\multicolumn{3}{c}{chrF}\\
\cmidrule(r){2-4}\cmidrule{5-7}
&
\multicolumn{1}{c}{T}&\multicolumn{1}{c}{MI}&\multicolumn{1}{c}{TMI}&
\multicolumn{1}{c}{T}&\multicolumn{1}{c}{MI}&\multicolumn{1}{c}{TMI}\\
\midrule}
\tablehead{\toprule
\multirow{2}{*}{Language}&\multicolumn{3}{c}{BLEU}&\multicolumn{3}{c}{chrF}\\
\cmidrule(r){2-4}\cmidrule{5-7}
&
\multicolumn{1}{c}{T}&\multicolumn{1}{c}{MI}&\multicolumn{1}{c}{TMI}&
\multicolumn{1}{c}{T}&\multicolumn{1}{c}{MI}&\multicolumn{1}{c}{TMI}\\
\midrule}
\tabletail{\midrule}
\tablelasttail{\\\bottomrule}
\begin{xtabular}{lrrrrrr}
Amharic & 0.00 & 0.10 & 0.13 & 0.23 & 9.70 & 10.59 \\
Arabic & 0.00 & 40.03 & 38.68 & 0.44 & 60.02 & 60.08 \\
Armenian & 0.00 & 0.22 & 0.16 & 0.30 & 12.22 & 12.14 \\
Assamese & 0.14 & 32.00 & 34.53 & 0.36 & 53.57 & 54.93 \\
Asturian & 1.29 & 36.49 & 35.08 & 24.29 & 60.08 & 58.55 \\
Azerbaijani & 0.23 & 1.71 & 2.15 & 14.78 & 18.81 & 18.49 \\
Belarusian & 0.00 & 3.06 & 2.75 & 0.62 & 22.39 & 22.69 \\
Bengali & 0.34 & 36.65 & 37.60 & 0.42 & 59.06 & 59.43 \\
Bosnian & 0.67 & 8.08 & 7.14 & 18.97 & 29.14 & 27.16 \\
Bulgarian & 0.36 & 17.15 & 15.83 & 1.19 & 42.10 & 39.51 \\
Burmese & 0.00 & 0.32 & 0.13 & 0.14 & 8.90 & 2.68 \\
Cantonese Chinese & 0.00 & 20.51 & 24.36 & 1.35 & 40.93 & 46.39 \\
Catalan & 3.60 & 49.63 & 49.47 & 27.98 & 71.20 & 71.10 \\
Cebuano & 1.61 & 3.88 & 3.85 & 20.21 & 22.63 & 22.79 \\
Croatian & 0.62 & 8.68 & 8.59 & 17.30 & 29.55 & 28.44 \\
Czech & 0.61 & 9.00 & 10.04 & 17.38 & 30.42 & 29.01 \\
Danish & 1.14 & 15.90 & 15.41 & 23.89 & 40.56 & 39.21 \\
Dutch & 1.45 & 18.48 & 17.04 & 24.92 & 45.34 & 42.89 \\
Estonian & 0.31 & 2.10 & 1.72 & 18.91 & 19.14 & 19.46 \\
Filipino & 1.29 & 3.40 & 4.01 & 19.66 & 22.15 & 22.58 \\
Finnish & 0.32 & 2.49 & 2.62 & 18.16 & 18.66 & 19.17 \\
French & 7.94 & 45.35 & 44.67 & 32.60 & 67.77 & 67.68 \\
Fula & 0.72 & 1.05 & 1.48 & 16.62 & 13.47 & 14.55 \\
Galician & 1.22 & 41.94 & 40.65 & 25.49 & 65.93 & 65.26 \\
Ganda & 1.18 & 18.44 & 17.60 & 16.68 & 36.82 & 36.11 \\
Georgian & 0.07 & 0.20 & 0.21 & 0.43 & 13.49 & 13.23 \\
German & 1.20 & 36.00 & 33.68 & 23.82 & 59.34 & 57.43 \\
Greek & 0.17 & 6.37 & 6.21 & 0.99 & 26.82 & 27.17 \\
Gujarati & 0.22 & 36.33 & 36.29 & 0.67 & 58.34 & 58.08 \\
Hausa & 0.43 & 1.05 & 1.26 & 15.21 & 12.84 & 15.20 \\
Hebrew & 0.13 & 1.29 & 1.37 & 1.72 & 16.41 & 17.12 \\
Hindi & 0.63 & 39.83 & 41.96 & 0.83 & 61.32 & 62.88 \\
Hungarian & 0.37 & 1.99 & 2.34 & 17.05 & 18.48 & 19.19 \\
Icelandic & 0.00 & 2.08 & 2.40 & 15.36 & 15.55 & 16.57 \\
Igbo & 1.11 & 15.93 & 16.52 & 15.25 & 34.57 & 35.07 \\
Indonesian & 1.82 & 45.64 & 45.57 & 21.43 & 66.67 & 66.50 \\
Irish & 0.31 & 0.57 & 0.74 & 16.96 & 14.26 & 15.72 \\
Italian & 0.78 & 31.57 & 31.09 & 25.77 & 59.67 & 58.60 \\
Japanese & 0.00 & 14.12 & 15.19 & 0.26 & 34.84 & 36.55 \\
Javanese & 0.61 & 8.37 & 8.62 & 19.03 & 29.17 & 29.15 \\
Kabuverdianu & 0.91 & 21.50 & 18.24 & 20.98 & 43.23 & 38.59 \\
Kamba & 1.34 & 4.32 & 3.55 & 16.28 & 19.15 & 18.70 \\
Kannada & 0.32 & 32.33 & 32.37 & 0.71 & 53.96 & 54.26 \\
Kazakh & 0.00 & 1.51 & 1.46 & 0.59 & 18.03 & 17.72 \\
Khmer & 0.01 & 1.02 & 0.62 & 0.96 & 13.27 & 13.12 \\
Korean & 0.00 & 3.88 & 3.16 & 0.46 & 21.65 & 20.84 \\
Kyrgyz & 0.13 & 1.10 & 1.03 & 0.55 & 16.63 & 16.69 \\
Lao & 0.10 & 1.26 & 0.66 & 1.36 & 11.41 & 6.71 \\
Latvian & 0.00 & 2.08 & 2.28 & 16.62 & 19.47 & 19.11 \\
Lingala & 0.94 & 21.06 & 20.01 & 16.97 & 41.34 & 39.68 \\
Lithuanian & 0.26 & 2.88 & 2.76 & 17.68 & 21.19 & 20.15 \\
Luo & 0.88 & 1.62 & 1.72 & 18.16 & 15.90 & 17.31 \\
Luxembourgish & 0.73 & 4.93 & 6.32 & 22.68 & 27.13 & 28.97 \\
Macedonian & 0.18 & 16.05 & 15.28 & 0.85 & 39.64 & 38.03 \\
Malay & 1.17 & 39.34 & 36.23 & 20.05 & 61.61 & 59.43 \\
Malayalam & 0.17 & 33.81 & 33.97 & 0.53 & 55.61 & 55.82 \\
Maltese & 0.86 & 3.79 & 4.31 & 21.14 & 22.09 & 24.00 \\
Mandarin Chinese & 0.00 & 25.39 & 28.15 & 1.67 & 47.25 & 50.67 \\
Maori & 1.16 & 1.02 & 1.47 & 17.30 & 12.23 & 15.04 \\
Marathi & 0.36 & 34.57 & 34.83 & 0.59 & 56.50 & 56.79 \\
Mongolian & 0.07 & 1.14 & 1.09 & 0.37 & 15.69 & 15.58 \\
Nepali & 0.53 & 38.47 & 39.51 & 0.84 & 59.55 & 60.40 \\
Northern-Sotho & 1.89 & 21.53 & 19.67 & 19.21 & 40.28 & 37.66 \\
Norwegian & 0.89 & 16.80 & 15.61 & 22.98 & 39.70 & 38.24 \\
Nyanja & 1.74 & 20.25 & 18.51 & 19.15 & 39.93 & 37.72 \\
Occitan & 0.74 & 37.59 & 36.38 & 25.82 & 61.36 & 59.98 \\
Oriya & 0.40 & 34.31 & 34.39 & 0.55 & 56.03 & 56.29 \\
Oromo & 0.00 & 0.00 & 0.00 & 14.63 & 11.12 & 15.17 \\
Pashto & 0.00 & 1.43 & 1.24 & 0.39 & 12.65 & 14.64 \\
Persian & 0.00 & 8.29 & 9.31 & 0.36 & 29.72 & 31.61 \\
Polish & 0.36 & 8.77 & 8.67 & 17.62 & 32.04 & 31.12 \\
Portuguese & 4.19 & 51.43 & 50.88 & 27.19 & 72.55 & 71.97 \\
Punjabi & 0.20 & 35.54 & 35.94 & 0.47 & 56.70 & 57.49 \\
Romanian & 0.90 & 23.08 & 19.88 & 24.80 & 48.30 & 44.61 \\
Russian & 0.65 & 26.62 & 24.75 & 2.26 & 51.51 & 49.17 \\
Serbian & 0.31 & 13.38 & 11.59 & 9.99 & 35.41 & 33.44 \\
Shona & 1.24 & 18.03 & 15.79 & 18.82 & 38.37 & 35.68 \\
Sindhi & 0.23 & 1.51 & 1.76 & 0.66 & 14.88 & 16.77 \\
Slovak & 0.45 & 6.69 & 6.71 & 17.76 & 27.78 & 26.39 \\
Slovenian & 0.31 & 3.39 & 3.31 & 17.98 & 22.74 & 22.55 \\
Somali & 0.43 & 0.80 & 0.89 & 15.08 & 12.28 & 14.65 \\
Sorani-Kurdish & 0.00 & 0.80 & 0.83 & 0.26 & 11.47 & 11.27 \\
Spanish & 1.30 & 38.84 & 38.14 & 25.48 & 63.95 & 63.67 \\
Swahili & 1.26 & 39.96 & 40.52 & 16.04 & 60.48 & 60.71 \\
Swedish & 0.86 & 18.38 & 18.09 & 23.47 & 41.82 & 40.74 \\
Tajik & 0.00 & 1.15 & 1.05 & 0.47 & 16.00 & 15.90 \\
Tamil & 1.43 & 31.29 & 32.42 & 1.52 & 52.13 & 54.22 \\
Telugu & 1.00 & 29.02 & 32.01 & 1.58 & 51.14 & 53.91 \\
Thai & 0.00 & 1.12 & 0.93 & 0.96 & 16.46 & 16.51 \\
Turkish & 0.34 & 3.84 & 3.94 & 17.18 & 21.28 & 21.73 \\
Ukrainian & 0.13 & 15.77 & 15.46 & 1.03 & 40.79 & 38.48 \\
Umbundu & 0.21 & 2.18 & 1.46 & 14.67 & 13.28 & 13.82 \\
Urdu & 1.68 & 32.62 & 31.91 & 2.20 & 53.99 & 53.91 \\
Uzbek & 0.11 & 0.75 & 0.79 & 17.11 & 14.21 & 17.17 \\
Vietnamese & 0.74 & 20.35 & 24.98 & 11.63 & 42.00 & 45.84 \\
Welsh & 0.95 & 1.41 & 1.60 & 17.99 & 16.89 & 18.12 \\
Wolof & 1.10 & 10.34 & 8.68 & 16.40 & 28.06 & 26.26 \\
Xhosa & 0.72 & 21.52 & 19.08 & 18.71 & 41.34 & 38.80 \\
Yoruba & 1.03 & 13.93 & 16.23 & 11.84 & 32.25 & 34.95 \\
Zulu & 0.50 & 23.34 & 20.13 & 17.82 & 42.55 & 40.54 \\
\midrule
Median & 0.43 & 11.86 & 10.81 & 15.70 & 32.15 & 32.52 \\
Average & 0.71 & 15.87 & 15.72 & 11.87 & 34.78 & 34.69
\end{xtabular}

\end{center}

\newpage

\section{Training Dataset}
\begin{center}
\footnotesize
\topcaption{}
\bottomcaption{}
\tablecaption{Training data breakdown by language and source dataset (CV stands for Common Voice).}
\tablefirsthead{\toprule
\multirow{2}{*}{Language}&\multicolumn{3}{c}{Utterances}&\multicolumn{3}{c}{Hours}\\
\cmidrule(r){2-4}\cmidrule{5-7}
&
\multicolumn{1}{c}{FLEURS}&\multicolumn{1}{c}{CV}&\multicolumn{1}{c}{Total}&
\multicolumn{1}{c}{FLEURS}&\multicolumn{1}{c}{CV}&\multicolumn{1}{c}{Total}\\
\midrule}
\tablehead{\toprule
\multirow{2}{*}{Language}&\multicolumn{3}{c}{Utterances}&\multicolumn{3}{c}{Hours}\\
\cmidrule(r){2-4}\cmidrule(r){5-7}
&
\multicolumn{1}{c}{FLEURS}&\multicolumn{1}{c}{CV}&\multicolumn{1}{c}{Total}&
\multicolumn{1}{c}{FLEURS}&\multicolumn{1}{c}{CV}&\multicolumn{1}{c}{Total}\\
\midrule}
\tabletail{\midrule}
\tablelasttail{\\\bottomrule}
\begin{xtabular}{lrrrrrr}
Abkhaz & - & 16,412 & 16,412 & - & 25.00 & 25.00 \\
Afrikaans & 1,025 & - & 1,025 & 3.58 & - & 3.58 \\
Amharic & 3,155 & - & 3,155 & 11.04 & - & 11.04 \\
Arabic & 2,098 & 21,948 & 24,046 & 6.02 & 25.00 & 31.02 \\
Armenian & 3,048 & 617 & 3,665 & 10.33 & 1.07 & 11.40 \\
Assamese & 2,776 & 831 & 3,607 & 10.35 & 1.34 & 11.69 \\
Asturian & 2,507 & 118 & 2,625 & 7.51 & 0.14 & 7.65 \\
Azerbaijani & 2,660 & 39 & 2,699 & 9.28 & 0.05 & 9.32 \\
Basaa & - & 763 & 763 & - & 0.93 & 0.93 \\
Bashkir & - & 20,836 & 20,836 & - & 25.00 & 25.00 \\
Basque & - & 10,904 & 10,904 & - & 15.92 & 15.92 \\
Belarusian & 2,410 & 18,347 & 20,757 & 9.31 & 25.00 & 34.31 \\
Bengali & 2,992 & 15,598 & 18,590 & 10.61 & 25.00 & 35.61 \\
Bosnian & 3,086 & - & 3,086 & 9.96 & - & 9.96 \\
Breton & - & 2,644 & 2,644 & - & 2.12 & 2.12 \\
Bulgarian & 2,966 & 3,212 & 6,178 & 9.45 & 4.65 & 14.10 \\
Burmese & 3,041 & - & 3,041 & 12.00 & - & 12.00 \\
Cantonese Chinese & 1,908 & 2,959 & 4,867 & 6.98 & 3.38 & 10.36 \\
Catalan & 2,294 & 16,188 & 18,482 & 7.39 & 25.00 & 32.39 \\
Cebuano & 3,242 & - & 3,242 & 12.00 & - & 12.00 \\
Chuvash & - & 1,538 & 1,538 & - & 2.06 & 2.06 \\
Croatian & 3,449 & - & 3,449 & 11.68 & - & 11.68 \\
Czech & 2,806 & 14,815 & 17,621 & 8.41 & 19.57 & 27.97 \\
Danish & 2,461 & 2,734 & 5,195 & 7.48 & 3.29 & 10.77 \\
Dhivehi & - & 2,682 & 2,682 & - & 3.81 & 3.81 \\
Dutch & 2,915 & 20,257 & 23,172 & 7.65 & 25.00 & 32.65 \\
English & 2,594 & 15,835 & 18,429 & 7.43 & 25.00 & 32.43 \\
Erzya & - & 1,241 & 1,241 & - & 1.97 & 1.97 \\
Esperanto & - & 14,503 & 14,503 & - & 25.00 & 25.00 \\
Estonian & 2,495 & 3,137 & 5,632 & 7.26 & 5.82 & 13.07 \\
Filipino & 1,868 & - & 1,868 & 7.57 & - & 7.57 \\
Finnish & 2,699 & 2,121 & 4,820 & 8.77 & 2.73 & 11.50 \\
French & 3,190 & 17,412 & 20,602 & 10.31 & 25.00 & 35.31 \\
Frisian & - & 3,799 & 3,799 & - & 5.21 & 5.21 \\
Fula & 3,136 & - & 3,136 & 12.89 & - & 12.89 \\
Galician & 2,172 & 5,021 & 7,193 & 6.67 & 6.36 & 13.03 \\
Ganda & 2,302 & - & 2,302 & 10.95 & - & 10.95 \\
Georgian & 1,478 & 3,944 & 5,422 & 4.96 & 6.27 & 11.22 \\
German & 2,984 & 15,766 & 18,750 & 8.99 & 25.00 & 33.99 \\
Greek & 3,210 & 1,919 & 5,129 & 10.01 & 2.08 & 12.09 \\
Guarani & - & 1,393 & 1,393 & - & 1.53 & 1.53 \\
Gujarati & 3,141 & - & 3,141 & 8.95 & - & 8.95 \\
Hakha Chin & - & 817 & 817 & - & 0.65 & 0.65 \\
Hausa & 3,171 & 1,930 & 5,101 & 12.77 & 2.28 & 15.05 \\
Hebrew & 3,235 & - & 3,235 & 9.42 & - & 9.42 \\
Hill Mari & - & 7,173 & 7,173 & - & 8.40 & 8.40 \\
Hindi & 2,114 & 4,437 & 6,551 & 6.61 & 5.23 & 11.84 \\
Hungarian & 3,091 & 7,744 & 10,835 & 9.27 & 10.86 & 20.13 \\
Icelandic & 924 & - & 924 & 2.83 & - & 2.83 \\
Igbo & 2,632 & 8 & 2,640 & 11.72 & 0.01 & 11.73 \\
Indonesian & 2,568 & 5,040 & 7,608 & 9.01 & 7.78 & 16.79 \\
Interlingua & - & 5,030 & 5,030 & - & 5.22 & 5.22 \\
Irish & 2,800 & 537 & 3,337 & 11.71 & 0.58 & 12.29 \\
Italian & 3,026 & 17,032 & 20,058 & 8.98 & 25.00 & 33.98 \\
Japanese & 2,291 & 7,211 & 9,502 & 7.42 & 9.94 & 17.36 \\
Javanese & 3,042 & - & 3,042 & 11.12 & - & 11.12 \\
Kabuverdianu & 2,694 & - & 2,694 & 10.33 & - & 10.33 \\
Kabyle & - & 26,356 & 26,356 & - & 25.00 & 25.00 \\
Kamba & 3,268 & - & 3,268 & 14.06 & - & 14.06 \\
Kannada & 2,270 & - & 2,270 & 8.16 & - & 8.16 \\
Kazakh & 3,186 & 453 & 3,639 & 11.68 & 0.63 & 12.31 \\
Khmer & 1,661 & - & 1,661 & 6.98 & - & 6.98 \\
Kinyarwanda & - & 17,733 & 17,733 & - & 25.00 & 25.00 \\
Korean & 2,304 & 94 & 2,398 & 7.92 & 0.16 & 8.08 \\
Kurmanji Kurdish & - & 4,426 & 4,426 & - & 4.88 & 4.88 \\
Kyrgyz & 2,816 & 1,787 & 4,603 & 9.31 & 2.32 & 11.63 \\
Lao & 1,793 & - & 1,793 & 7.20 & - & 7.20 \\
Latvian & 2,105 & 2,734 & 4,839 & 6.49 & 2.38 & 8.87 \\
Lingala & 2,991 & - & 2,991 & 14.60 & - & 14.60 \\
Lithuanian & 2,929 & 5,196 & 8,125 & 9.68 & 7.12 & 16.80 \\
Luganda & - & 15,037 & 15,037 & - & 25.00 & 25.00 \\
Luo & 2,294 & - & 2,294 & 9.14 & - & 9.14 \\
Luxembourgish & 2,486 & - & 2,486 & 8.33 & - & 8.33 \\
Macedonian & 2,333 & 115 & 2,448 & 6.77 & 0.16 & 6.93 \\
Malay & 2,658 & - & 2,658 & 9.48 & - & 9.48 \\
Malayalam & 3,031 & 459 & 3,490 & 9.95 & 0.54 & 10.50 \\
Maltese & 2,891 & 1,944 & 4,835 & 9.89 & 2.42 & 12.30 \\
Mandarin Chinese & 3,239 & 6,655 & 9,894 & 9.68 & 6.00 & 15.68 \\
Maori & 2,940 & - & 2,940 & 15.10 & - & 15.10 \\
Marathi & 3,250 & 2,238 & 5,488 & 11.78 & 3.71 & 15.49 \\
Meadow Mari & - & 19,365 & 19,365 & - & 25.00 & 25.00 \\
Moksha & - & 173 & 173 & - & 0.26 & 0.26 \\
Mongolian & 2,971 & 2,149 & 5,120 & 10.50 & 3.07 & 13.57 \\
Nepali & 3,322 & 167 & 3,489 & 11.18 & 0.18 & 11.36 \\
Northern-Sotho & 1,570 & - & 1,570 & 8.69 & - & 8.69 \\
Norwegian & 3,156 & 314 & 3,470 & 10.82 & 0.38 & 11.20 \\
Nyanja & 2,649 & - & 2,649 & 10.40 & - & 10.40 \\
Occitan & 3,295 & 41 & 3,336 & 13.45 & 0.06 & 13.52 \\
Odia & - & 482 & 482 & - & 0.68 & 0.68 \\
Oriya & 1,079 & - & 1,079 & 3.42 & - & 3.42 \\
Oromo & 1,688 & - & 1,688 & 6.51 & - & 6.51 \\
Pashto & 2,494 & - & 2,494 & 8.72 & - & 8.72 \\
Persian & 3,077 & 23,479 & 26,556 & 11.86 & 25.00 & 36.86 \\
Polish & 2,839 & 16,916 & 19,755 & 9.17 & 24.80 & 33.97 \\
Portuguese & 2,782 & 19,282 & 22,064 & 10.09 & 21.94 & 32.04 \\
Punjabi & 1,917 & 695 & 2,612 & 6.32 & 1.02 & 7.34 \\
Quechua Chanka & - & 1 & 1 & - & 0.00 & 0.00 \\
Romanian & 2,887 & 5,113 & 8,000 & 10.10 & 5.65 & 15.75 \\
Romansh Sursilvan & - & 1,552 & 1,552 & - & 2.43 & 2.43 \\
Romansh Vallader & - & 671 & 671 & - & 1.18 & 1.18 \\
Russian & 2,559 & 17,444 & 20,003 & 8.03 & 25.00 & 33.04 \\
Sakha & - & 1,594 & 1,594 & - & 2.63 & 2.63 \\
Santali (Ol Chiki) & - & 279 & 279 & - & 0.37 & 0.37 \\
Saraiki & - & 1,256 & 1,256 & - & 1.22 & 1.22 \\
Sardinian & - & 458 & 458 & - & 0.53 & 0.53 \\
Serbian & 2,919 & 1,380 & 4,299 & 10.44 & 1.05 & 11.49 \\
Shona & 2,442 & - & 2,442 & 9.78 & - & 9.78 \\
Sindhi & 3,420 & - & 3,420 & 12.11 & - & 12.11 \\
Slovak & 1,955 & 2,967 & 4,922 & 5.86 & 3.11 & 8.97 \\
Slovenian & 2,504 & 1,461 & 3,965 & 7.69 & 1.43 & 9.13 \\
Somali & 3,051 & - & 3,051 & 12.31 & - & 12.31 \\
Sorani-Kurdish & 3,028 & 7,010 & 10,038 & 10.34 & 8.03 & 18.37 \\
Sorbian, Upper & - & 808 & 808 & - & 1.48 & 1.48 \\
Spanish & 2,795 & 17,155 & 19,950 & 8.80 & 25.00 & 33.80 \\
Swahili & 2,993 & 16,481 & 19,474 & 12.71 & 25.00 & 37.71 \\
Swedish & 2,372 & 7,421 & 9,793 & 8.25 & 8.20 & 16.44 \\
Taiwanese (Minnan) & - & 1,646 & 1,646 & - & 1.20 & 1.20 \\
Tajik & 2,289 & - & 2,289 & 8.53 & - & 8.53 \\
Tamil & 2,351 & 13,775 & 16,126 & 8.53 & 25.00 & 33.53 \\
Tatar & - & 9,565 & 9,565 & - & 10.11 & 10.11 \\
Telugu & 2,296 & - & 2,296 & 7.87 & - & 7.87 \\
Thai & 2,596 & 21,797 & 24,393 & 8.44 & 25.00 & 33.44 \\
Tigre & - & 10 & 10 & - & 0.01 & 0.01 \\
Tigrinya & - & 10 & 10 & - & 0.02 & 0.02 \\
Toki Pona & - & 2,450 & 2,450 & - & 2.34 & 2.34 \\
Turkish & 2,521 & 26,036 & 28,557 & 8.27 & 25.00 & 33.27 \\
Twi & - & 12 & 12 & - & 0.01 & 0.01 \\
Ukrainian & 2,805 & 15,749 & 18,554 & 9.00 & 18.64 & 27.64 \\
Umbundu & 1,149 & - & 1,149 & 6.44 & - & 6.44 \\
Urdu & 2,101 & 4,130 & 6,231 & 6.96 & 4.98 & 11.94 \\
Uyghur & - & 4,421 & 4,421 & - & 7.43 & 7.43 \\
Uzbek & 2,939 & 22,042 & 24,981 & 10.05 & 25.00 & 35.05 \\
Vietnamese & 2,988 & 2,475 & 5,463 & 9.03 & 3.12 & 12.15 \\
Votic & - & 96 & 96 & - & 0.11 & 0.11 \\
Welsh & 3,354 & 7,769 & 11,123 & 11.56 & 11.06 & 22.62 \\
Wolof & 2,263 & - & 2,263 & 8.58 & - & 8.58 \\
Xhosa & 3,430 & - & 3,430 & 13.01 & - & 13.01 \\
Yoruba & 2,293 & 39 & 2,332 & 9.60 & 0.07 & 9.67 \\
Zulu & 2,720 & - & 2,720 & 13.54 & - & 13.54 \\
\midrule
Median & 2,748 & 2,963 & 3,470 & 9.27 & 3.76 & 11.36 \\
Total & 268,000 & 725,660 & 993,660 & 950.09 & 954.98 & 1905.07
\end{xtabular}

\end{center}

\end{document}